\def\year{2018}\relax
\documentclass[letterpaper]{article} 
\usepackage{custom_aaai18}  
\usepackage{times}  
\usepackage{helvet}  
\usepackage{courier}  
\usepackage{url}  
\usepackage{graphicx}  
\usepackage{amssymb} 
\usepackage{amsmath} 
\usepackage{booktabs} 
\usepackage{subcaption} 
\usepackage{minipage}

\newcommand{\R}{\boldsymbol R}
\newcommand{\E}{\boldsymbol E}
\newcommand{\U}{\boldsymbol U}
\newcommand{\Ui}{\boldsymbol{U_i}}
\newcommand{\V}{\boldsymbol V}
\newcommand{\Vj}{\boldsymbol{V_j}}

\newcommand{\btheta}{\boldsymbol \theta}

\newcommand{\sumOmega}{\sum_{(i,j) \in \Omega}}

\begin{document}
%
\title{Prior and Likelihood Choices for Bayesian Matrix Factorisation on Small Datasets}
\author{Thomas Brouwer \and Pietro Li\'{o} \\
Computer Laboratory, University of Cambridge, United Kingdom \\
}
\maketitle
\begin{abstract}
	In this paper, we study the effects of different prior and likelihood choices for Bayesian matrix factorisation, focusing on small datasets. These choices can greatly influence the predictive performance of the methods. We identify four groups of approaches: Gaussian-likelihood with real-valued priors, nonnegative priors, semi-nonnegative models, and finally Poisson-likelihood approaches. For each group we review several models from the literature, considering sixteen in total, and discuss the relations between different priors and matrix norms. We extensively compare these methods on eight real-world datasets across three application areas, giving both inter- and intra-group comparisons. We measure convergence runtime speed, cross-validation performance, sparse and noisy prediction performance, and model selection robustness. We offer several insights into the trade-offs between prior and likelihood choices for Bayesian matrix factorisation on small datasets---such as that Poisson models give poor predictions, and that nonnegative models are more constrained than real-valued ones.
\end{abstract}

\section{Introduction}
	Matrix factorisation methods have become very popular in recent years, and used for many applications such as collaborative filtering \cite{Mnih2008,Chen2009} and bioinformatics \cite{Gonen2012,Brouwer2017a}. Given a matrix relating two entity types, such as movies and users, matrix factorisation decomposes that matrix into two smaller so-called factor matrices, such that their product approximates the original one. Matrix factorisation is often used for predicting missing values in the datasets, and analysing the resulting factor values to identify biclusters or features.
	
	Most models can be categorised as being either non-probabilistic, such as the popular models by \cite{Lee2000}, or Bayesian. The former seek to minimise an error function (such as the squared error) between the original matrix and the approximation. In contrast, Bayesian variants treat the two smaller matrices as random variables, place prior distributions over them, and find the posterior distribution over their values after observing the data. A likelihood function, usually Gaussian, is used to capture noise in the dataset. Previous work \cite{Brouwer2017a} has demonstrated that Bayesian variants are much better for predictive tasks than non-probabilistic versions, which tend to overfit to noise and sparsity. 
	
	Matrix factorisation techniques can also be grouped by their constraints on the values in the factor matrices. Firstly, many approaches place no constraints, using real-valued factor matrices (commonly done in the Bayesian literature \cite{Salakhutdinov2008,Gonen2012}). Instead, we could constrain them to be nonnegative (as is popular in the non-probabilistic literature \cite{Lee2000,Tan2013}), limiting its applicability to nonnegative datasets, but making it easier to interpret the factors and potentially also making the method more robust to overfitting. Thirdly, semi-nonnegative variants constrain one factor matrix to be nonnegative, leaving the other real-valued \cite{FeiWangTaoLi2008,Ding2010}. Finally, some versions work only on count data.
	
	In the Bayesian setting, the first three groups of methods all generally use a Gaussian likelihood for noise, and place either real-valued or nonnegative priors over the matrices. For the former, Gaussian is a common choice \cite{Salakhutdinov2008,Gonen2012,Virtanen2011,Virtanen2012}, and for the latter options include the exponential distributions \cite{Schmidt2009}. The fourth group uses a Poisson likelihood to capture count data \cite{Gopalan2014,Gopalan2015,Hu2015}. These models are often extended by using complicated hierarchical prior structures over the factor matrices, giving additional behaviour (such as automatic model selection).
	
	This paper offers the first systematic comparison between different Bayesian variants of matrix factorisation. Similar comparisons have been provided in other fields, such as for the regression parameter in Bayesian model averaging \cite{Ley2009,Eicher2011}, which demonstrated that the choice of prior can greatly influence the predictive performance of these models. However, a similar study for Bayesian matrix factorisation is still missing. More strikingly, many papers that introduce new matrix factorisation models do not provide a thorough comparison with competing approaches, or popular non-probabilistic ones such as \cite{Lee2000}---for example, the seminal paper by \cite{Salakhutdinov2008} compares their approach with only one other matrix factorisation method; although \cite{Gopalan2015} compares with three others.
	
	We give an overview of the different approaches that can be found in the literature, including hierarchical priors, and then study the effects of these different Bayesian prior and likelihood choices. 
	We aim to make general statements about the behaviour of the four different groups of methods on small real-world datasets (up to a million observed entries), by considering eight datasets across three different applications---four drug effectiveness datasets, two collaborative filtering datasets, and two methylation expression datasets. Our experiments consider convergence speed, cross-validation performance, sparse and noisy prediction performance, and model selection effectiveness. 
	This study offers novel insights into the differences between the four approaches, and the effects of popular hierarchical priors. 
	
	
	We note that there is a rich literature of Bayesian nonparametric matrix factorisation models, which learn the size of the factor matrices automatically. However, these models often require complex inference approaches to find good solutions, and hence their predictive performance is more determined by the inference method than the precise model choices (such as likelihood and prior). In this paper we therefore focus on parametric matrix factorisation models, to isolate the effects of likelihood and prior choices.
	
	Finally, we acknowledge that the models we study were generally introduced for a specific application domain, and that this makes it hard to make general statements about the behaviour of these methods on different datasets. However, we believe that it is essential to provide a cross-application comparison of the different approaches, as this teaches us valuable lessons for the applications studied, and they are likely to apply to different areas as well. The lack of other studies exploring the trade-offs between likelihood and prior choices for Bayesian matrix factorisation make this a novel and essential study.

\section{Bayesian Matrix Factorisation}
	In this section we introduce the different matrix factorisation models that we study. Formally, the problem of matrix factorisation can be defined as follows. Given an observed matrix $ \R \in \mathbb{R}^{I \times J} $, we want to find two smaller matrices $ \U \in \mathbb{R}^{I \times K} $ and $ \V \in \mathbb{R}^{J \times K} $, each with $K$ so-called factors (columns), to solve $ \R = \U \V^T + \E $, where noise is captured by matrix $ \E \in \mathbb{R}^{I \times J} $. Some entries in $\R$ may be unobserved, as given by the set $ \Omega = \left\{ (i,j) \text{ $ \vert $ $ R_{ij} $ is observed} \right\} $. These entries can then be predicted by $\U \V^T$.
	
	In the Bayesian treatment of matrix factorisation, we express a likelihood function for the observed data that captures noise (such as Gaussian or Poisson). We treat the latent matrices as random variables, placing prior distributions over them. A Bayesian solution for matrix factorisation can then be found by inferring the posterior distribution $p(\btheta|D)$ over the latent variables $\btheta$ ($\U$, $\V$, and any additional random variables in our model), given the observed data $ D = \lbrace R_{ij} \rbrace_{i,j \in \Omega} $. This posterior distribution is often intractable to compute exactly, but several methods exist to approximate it (see Section \ref{Inference}). 
	
	In Section \ref{Models} we introduce a wide range of models from the literature, and categorise them into four groups. The model names are highlighted in bold in the text.
	
	\subsection{Probability Distributions}
		We introduce all notation and probability distributions in the paper below. 
		
		$\text{diag}(\boldsymbol \lambda^{-1})$ is a diagonal matrix with entries $\lambda_1^{-1}, .., \lambda_K^{-1}$ on the diagonal.
		
		$ \mathcal{N} (x|\mu,\tau^{-1}) = \tau^{\frac{1}{2}} (2\pi)^{-\frac{1}{2}} \exp \left\{ -\frac{\tau}{2} (x - \mu)^2 \right\} $ is a Gaussian distribution with precision $ \tau $. 
		
		$ \mathcal{N} (\boldsymbol x|\boldsymbol \mu,\boldsymbol \Sigma) = \vert \boldsymbol \Sigma \vert^{-\frac{1}{2}} (2\pi)^{-\frac{K}{2}} \exp \left\{ -\frac{1}{2} (\boldsymbol x - \boldsymbol \mu)^T \boldsymbol \Sigma^{-1} (\boldsymbol x - \boldsymbol \mu) \right\} $ is a $K$-dimensional multivariate Gaussian distribution.
		
		$ \mathcal{G} (\tau | \alpha_{\tau}, \beta_{\tau} ) = \frac{{\beta_{\tau}}^{\alpha_{\tau}}}{\Gamma(\alpha_{\tau})} x^{\alpha_{\tau} -1} e^{- \beta_{\tau} x} $ is a Gamma distribution, where $ \Gamma(x) = \int_{0}^{\infty} x^{t-1} e^{-x} dt $ is the gamma function. 
		
		$ \mathcal{NIW} ( \boldsymbol{\mu}, \boldsymbol{\Sigma} | \boldsymbol{\mu_0}, \beta_0, \nu_0, \boldsymbol{W_0} ) = \mathcal{N}(\boldsymbol \mu | \boldsymbol{\mu_0}, \frac{1}{\beta_0} \text{\textbf I} ) \mathcal{W}^{-1} ( \boldsymbol \Sigma | \nu_0, \boldsymbol{W_0} ) $ is a normal-inverse Wishart distribution, where $\mathcal{W}^{-1} ( \boldsymbol \Sigma | \nu_0, \boldsymbol{W_0} ) $ is an inverse Wishart distribution, and $\textbf I$ the identity matrix.
		
		$\mathcal{L} ( x | \mu, \rho ) = \frac{1}{2 \rho} \exp \left\{ -\frac{ |x - \mu| }{\rho} \right\} $ is a Laplace distribution.
		
		$ \mathcal{IG}(x|\mu, \lambda) = \frac{ \lambda }{2 \pi x^3} \exp \left\{ - \frac{\lambda (x - \mu)^2}{2 \mu^2 x} \right\} $ is an inverse Gaussian.
		
		$ \mathcal{E} ( x | \lambda ) = \lambda \exp \left\{ - \lambda x \right\} u(x) $ is an exponential distribution, where $u(x)$ is the unit step function.
		\begin{equation*}
			\mathcal{TN} ( x | \mu, \tau ) = \left\{
			\begin{array}{ll}
			\displaystyle \frac{ \sqrt{ \frac{\tau}{2\pi} } \exp \left\{ -\frac{\tau}{2} (x - \mu)^2 \right\} }{ 1 - \Phi ( - \mu \sqrt{\tau} )}  & \mbox{if } x \geq 0 \\
			0 & \mbox{if } x < 0
			\end{array}
			\right.
		\end{equation*}
		is a truncated normal: a normal distribution with zero density below $ x = 0 $ and renormalised to integrate to one. $ \Phi(\cdot) $ is the cumulative distribution function of $ \mathcal{N}(0,1) $.

	\begin{table*}[t]
		\caption{Overview of the Bayesian matrix factorisation models.} \label{overview_bmf_models}
		\centering
		\begin{tabular}{llllll}
			\toprule
			Category & Name & Likelihood \hspace{2pt} & Prior $\U$ & Prior $\V$ & Hierarchical prior \\
			\midrule
			Real-valued & GGG & $\mathcal{N} (R_{ij} | \Ui \Vj, \tau^{-1} )$ & $\mathcal{N} ( \Ui | \boldsymbol 0, \lambda^{-1} \text{\textbf I} )$ & $\mathcal{N} ( \Vj | \boldsymbol 0, \lambda^{-1} \text{\textbf I} )$ & - \\
			& GGGU & $\mathcal{N} (R_{ij} | \Ui \Vj, \tau^{-1} )$ & $\mathcal{N} ( \Ui | \boldsymbol 0, \lambda^{-1} \text{\textbf I} )$ & $\mathcal{N} ( \Vj | \boldsymbol 0, \lambda^{-1} \text{\textbf I} )$ & - \\
			& GGGA & $\mathcal{N} (R_{ij} | \Ui \Vj, \tau^{-1} )$ & $\mathcal{N} ( \Ui | \boldsymbol 0, \text{diag}(\boldsymbol \lambda^{-1}) )$ & $\mathcal{N} ( \Vj| \boldsymbol 0, \text{diag}(\boldsymbol \lambda^{-1}) )$ & $\lambda_k \sim \mathcal{G} ( \alpha_0, \beta_0 )$ \\
			& GGGW & $\mathcal{N} (R_{ij} | \Ui \Vj, \tau^{-1} )$ & $\mathcal{N} ( \Ui | \boldsymbol{\mu_U}, \boldsymbol{\Sigma_U})$ & $\mathcal{N} ( \Vj | \boldsymbol{\mu_V}, \boldsymbol{\Sigma_V})$ & $(\boldsymbol{\mu_U}, \boldsymbol{\Sigma_U}) \text{ and } (\boldsymbol{\mu_V}, \boldsymbol{\Sigma_V}) $ \\
			& & & & & $\sim \mathcal{NIW} ( \boldsymbol{\mu_0}, \beta_0, \nu_0, \boldsymbol{W_0} )$ \\
			& GLL & $\mathcal{N} (R_{ij} | \Ui \Vj, \tau^{-1} )$ & $\mathcal{L} ( U_{ik} | 0, \eta ) $ & $\mathcal{L} ( V_{jk} | 0, \eta ) $ & - \\
			& GLLI & $\mathcal{N} (R_{ij} | \Ui \Vj, \tau^{-1} )$ & $\mathcal{L} ( U_{ik} | 0, \eta^U_{ik} ) $ & $\mathcal{L} ( V_{jk} | 0, \eta^V_{jk} ) $ & $\eta^U_{ik} \text{ and } \eta^V_{jk} \sim \mathcal{IG}(\mu, \lambda) $ \\
			& GVG & $\mathcal{N} (R_{ij} | \Ui \Vj, \tau^{-1} )$ & $p(\U) \propto $ & $\mathcal{N} ( \Vj | \boldsymbol 0, \lambda^{-1} \text{\textbf I} )$ & - \\
			& & & $\exp \lbrace - \gamma \det (\U^T \U) \rbrace$ & & \\
			\midrule
			Nonnegative & GEE & $\mathcal{N} (R_{ij} | \Ui \Vj, \tau^{-1} )$ & $\mathcal{E} ( U_{ik} | \lambda )$ & $\mathcal{E} ( V_{jk} | \lambda )$ & - \\
			& GEEA & $\mathcal{N} (R_{ij} | \Ui \Vj, \tau^{-1} )$ & $\mathcal{E} ( U_{ik} | \lambda_k )$ & $\mathcal{E} ( V_{jk} | \lambda_k )$ & $\lambda_k \sim \mathcal{G} ( \alpha_0, \beta_0 )$ \\
			& GTT & $\mathcal{N} (R_{ij} | \Ui \Vj, \tau^{-1} )$ & $\mathcal{TN} ( U_{ik} | \mu_U, \tau_U )$ & $\mathcal{TN} ( V_{jk} | \mu_V, \tau_V )$ & - \\
			& GTTN & $\mathcal{N} (R_{ij} | \Ui \Vj, \tau^{-1} )$ & $\mathcal{TN} ( U_{ik} | \mu_U, \tau_U )$ & $\mathcal{TN} ( V_{jk} | \mu_V, \tau_V )$ & $p(\mu^U_{ik}, \tau^U_{ik} | \mu_{\mu}, \tau_{\mu}, a, b) \propto $ \\
			& & & & & $\frac{1}{\sqrt{\tau^U_{ik}}} \left( 1 - \Phi ( - \mu^U_{ik} \sqrt{\tau^U_{ik}} ) \right)$ \\
			& & & & & $\mathcal{N} (\mu^U_{ik} | \mu_{\mu}, \tau_{\mu}^{-1} ) \mathcal{G} (\tau^U_{ik} | a, b)$ \\
			& G$\text{L}^2_1 $ & $\mathcal{N} (R_{ij} | \Ui \Vj, \tau^{-1} )$ & $p(\U) \propto \exp $ & $p(\V) \propto \exp $ & - \\
			& & & $ \lbrace -\frac{\lambda}{2} \sum_i ( \sum_k U_{ik} )^2 \rbrace $ & $ \lbrace -\frac{\lambda}{2} \sum_j ( \sum_k V_{jk} )^2 \rbrace $ & \\
			& & & with $U_{ik} \geq 0$ & with $V_{jk} \geq 0$ & \\
			\midrule
			Semi- & GEG & $\mathcal{N} (R_{ij} | \Ui \Vj, \tau^{-1} )$ & $\mathcal{E} ( U_{ik} | \lambda )$ & $\mathcal{N} ( \Vj | \boldsymbol 0, \lambda^{-1} \text{\textbf I} )$ & - \\
			nonnegative & GVnG & $\mathcal{N} (R_{ij} | \Ui \Vj, \tau^{-1} )$ & GVG with $U_{ik} \geq 0$ & $\mathcal{N} ( \Vj | \boldsymbol 0, \lambda^{-1} \text{\textbf I} )$ & - \\
			\midrule
			Poisson & PGG & $\mathcal{P} (R_{ij} | \Ui \Vj )$ & $\mathcal{G} ( U_{ik} | a, b )$ & $\mathcal{G} ( V_{jk} | a, b )$ & - \\
			& PGGG & $\mathcal{P} (R_{ij} | \Ui \Vj )$ & $\mathcal{G} ( U_{ik} | a, h^U_i )$ & $\mathcal{G} ( V_{jk} | a, h^V_j )$ & $h^U_i \text{ and } h^V_j \sim \mathcal{G} (a', \frac{a'}{b'})$ \\
			\bottomrule
		\end{tabular}
	\end{table*}

	\subsection{Models} \label{Models}
		There are three types of choices we make that determine the type of matrix factorisation model we use: the likelihood function, the priors we place over the factor matrices $\U$ and $\V$, and whether we use any further hierarchical priors. 
		We have identified four different groups of Bayesian matrix factorisation approaches based on these choices: Gaussian-likelihood with real-valued priors, nonnegative priors (constraining the matrices $\U, \V$ to be nonnegative), semi-nonnegative models (constraining one of the two factor matrices to be nonnegative), and finally Poisson-likelihood approaches. 
		Models within each group use different priors and hierarchical priors, and many choices can be found in the literature. In this paper we consider a total of sixteen models, as summarised in Table \ref{overview_bmf_models}. We have focused on fully conjugate models (meaning the prior and likelihood are in the same family of distributions) to ensure inference for each model is guaranteed to work well, so that all performance differences in Section \ref{Experiments} come entirely from the choice of likelihood and priors.
		
		The first three groups all use a Gaussian likelihood for noise, by assuming each value in $\R$ comes from the product of $\U$ and $\V$, $R_{ij} \sim \mathcal{N} (R_{ij} | \Ui \Vj, \tau^{-1} )$, with Gaussian noise added of precision $\tau$, for which we use a Gamma prior $\mathcal{G} (\tau | \alpha_{\tau}, \beta_{\tau} )$. 
		The last group instead opt for a Poisson likelihood, $ R_{ij} \sim \mathcal{P} (R_{ij} | \Ui \Vj )$. This only works for nonnegative count data, with $ \R \in \mathbb{N}^{I \times J} $, but has been studied extensively in the literature due to the popularity and prevalence of datasets like the Netflix Challenge. 
		
		\paragraph{Real-valued matrix factorisation}
			The most common approach is to use independent zero-mean Gaussian priors for $\U, \V$ \cite{Salakhutdinov2008,Gonen2012,Virtanen2011,Virtanen2012}, which gives rise to the \textbf{GGG} model. The \textbf{GGGU} model is identical but uses a univariate posterior for inference (see supplementary materials).
			
			The first hierarchical model (\textbf{GGGA}) uses the Bayesian automatic relevance determination (ARD) prior, which helps with model selection. The main idea is to replace the $\lambda$ hyperparameter by a factor-specific variable $\lambda_k$, which has a further Gamma prior. This causes all entries in columns of $\U$ and $\V$ to go further to zero if only a few values in that column are high, effectively making the factor inactive. This prior has been used for real-valued \cite{Virtanen2011,Virtanen2012} and nonnegative matrix factorisation \cite{Tan2013}.
			
			Another hierarchical model (\textbf{GGGW}) was introduced in the seminal paper of \cite{Salakhutdinov2008}. Instead of assuming independence of each entry in $\U, \V$, we assume each row of $\U$ comes from a multivariate Gaussian with row mean $\boldsymbol{\mu_U}$ and covariance $\boldsymbol{\Sigma_U}$, and similarly for $\V$. We then place a further Normal-Inverse Wishart prior over these parameters. 
			
			An alternative to the Gaussian prior is to use the Laplace distribution \cite{Jing2015}, which has a much more pointy distribution than Gaussian around $x=0$. This leads to more sparse solutions, as more factors are set to low values. The basic model (\textbf{GLL}) can be extended with a hierarchical Inverse Gaussian prior over the $\eta$ parameter (\textbf{GLLI}), which they claim helps with variable selection.
			
			The final model (\textbf{GVG}) was introduced by \cite{Arngren2011}. They used a volume prior for the $\U$ matrix, with density $p(\U) \propto \exp \lbrace - \gamma \det (\U^T \U) \rbrace $. The $\gamma$ hyperparameter determines the strength of the volume penalty (higher means stronger prior).
			
		\paragraph{Nonnegative matrix factorisation}
			These models all place nonnegative prior distributions over entries in $\U$ and $\V$, and as a result can only deal with nonnegative datasets. 
			
			\cite{Schmidt2009} introduced a model using exponential priors over the factor matrices (\textbf{GEE}). This model can also be extended with ARD \cite{Brouwer2017b} (\textbf{GEEA}). Another option is to use the truncated normal distribution (\textbf{GTT}), which can also be extended by placing a hierarchical prior over the mean and precision $\mu_U, \tau_U, \mu_V, \tau_V$ (\textbf{GTTN}), as done by \cite{MikkelN.Schmidt2009}. This nontrivial prior cannot be sampled from directly, but will be useful for inference.
			
			Finally, we can use a prior inspired by the $\text{L}^2_1$ norm for both $\U$ and $\V$ (\textbf{G$\boldsymbol{\text{L}^2_1}$}), as we will discuss in Section \ref{PriorsNorms}.
			
		\paragraph{Semi-nonnegative matrix factorisation} 
			Instead of forcing nonnegativity on both factor matrices, we could place this constraint on only one, as was done in \cite{FeiWangTaoLi2008,Ding2010}. In the Bayesian setting we place a real-valued prior over one matrix, and a nonnegative prior over the other. The major advantage is that we can handle real-valued datasets, while still enforcing some nonnegativity. However, we will see in Section \ref{Experiments} that its performance is identical to the real-valued approaches.
			
			Firstly we can use an exponential prior for entries in $\U$, and a Gaussian for $\V$, effectively combining the GGG and GEE models into one (\textbf{GEG}). 
			Another semi-nonnegative model (\textbf{GVnG}) comes from constraining the volume prior in the GVG model to also be nonnegative: $p(\U) = 0$ if any $U_{ik} < 0$.
			
		\paragraph{Poisson likelihood}
			The standard Poisson matrix factorisation model (\textbf{PGG}) uses independent Gamma priors over the entries in $\U$ and $\V$, with hyperparameters $a, b$ \cite{Gopalan2014,Gopalan2015,Hu2015}. 
			This model can also be extended with a hierarchical prior (\textbf{PGGG}), by replacing $b$ with $h^U_i, h^V_j$ and placing a further Gamma prior over these parameters \cite{Gopalan2015}.

\section{Priors and Norms} \label{PriorsNorms}
	The prior distributions in Bayesian models act as a regulariser that prevents us from overfitting to the data, preventing poor predictive performance. We can write out the expression of the log posterior of the parameters, which for a Gaussian likelihood and no hierarchical priors becomes
	\begin{alignat*}{1}
		& \log p(\btheta|D) = \log p(D|\btheta) + \log p(\btheta) + C_1 \\
		&\hspace{25pt} = \sumOmega \log p(R_{ij}|\Ui \Vj, \tau^{-1}) + \log p(\U, \V) + C_2 \\
		&\hspace{25pt} = - \frac{\tau}{2} \sumOmega ( R_{ij} - \Ui \Vj)^2 + \log p(\U, \V) + C_3
	\end{alignat*}
	for some constants $C_i$. Note that this last expression is simply the negative Frobenius norm (squared error) of the training fit, plus a regularisation term over the matrices $\U, \V$. This training error is frequently used in the nonprobabilistic matrix factorisation literature \cite{Lee2000,Pauca2004,Pauca2006}, where different regularisation terms are used. 
	These are often based on row-wise \textbf{matrix norms}, such as
	\begin{alignat*}{2}
		& \text{L}_1 = \sum_{i=1}^I \sum_{k=1}^K U_{ik}
		\quad\quad && \text{L}_2 = \sum_{i=1}^I \sqrt{\sum_{k=1}^K U_{ik}} \\
		& \text{L}^2_1 = \sum_{i=1}^I (\sum_{k=1}^K U_{ik})^2
		\quad\quad && \text{L}^2_2 = \sum_{i=1}^I \sum_{k=1}^K U_{ik}^2
	\end{alignat*}
	This offers some interesting insights: the $\text{L}^2_2$ norm is equivalent to an independent Gaussian prior (GGG), due to the square in the exponential of the Gaussian prior; the $\text{L}_1$ norm is equivalent to a Laplace prior distribution (GLL); if we constrain $\U, \V$ to be nonnegative then the $\text{L}_1$ norm is equivalent to an exponential prior distribution (GEE); and finally, the $\text{L}^2_1$ norm can be formulated as a nonnegative prior distribution, which we use for the $\text{GL}^2_1$ model (see Table \ref{overview_bmf_models}).
	
	In other words, the type of priors chosen for Bayesian matrix factorisation determine the type of regularisation that we add to the model. Additionally, we can use hierarchical priors to model further desired behaviour (such as ARD).

\section{Model Discussion} 
	\subsection{Inference} \label{Inference}
		In this paper we use Gibbs sampling (see Section \ref{Inference}), because it tends to be very accurate at finding the true posterior \cite{Brouwer2017b}, but other methods like variational Bayesian inference are also possible. The Gibbs sampling algorithms, together with their time complexities, are given in the supplementary materials.
		
	\subsection{Hyperparameters} \label{Hyperparameters}
		\begin{figure}[t]
			\centering
			\includegraphics[width=\columnwidth]{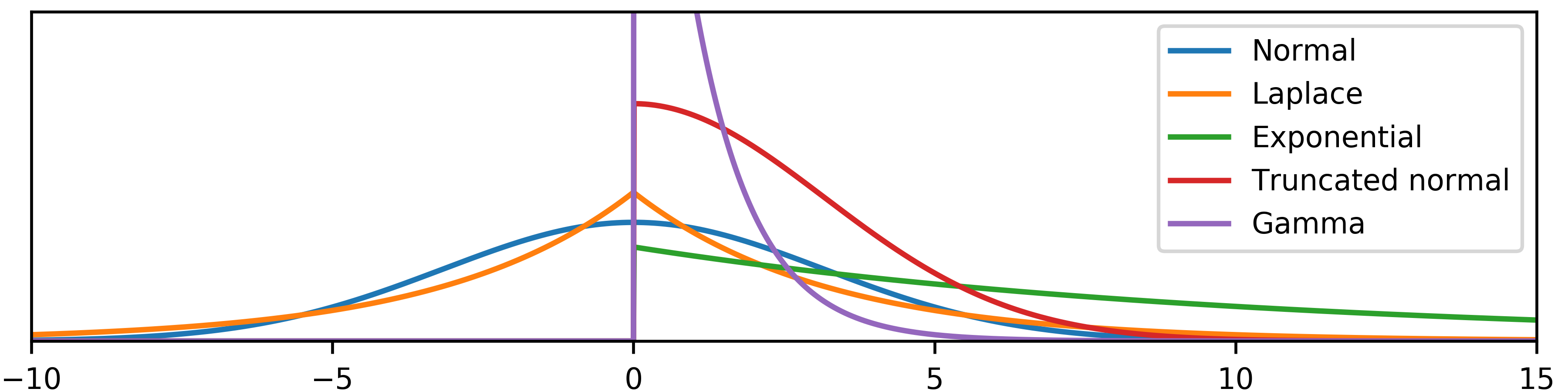}
			\caption{Plots of the prior distributions with hyperparameters from Section \ref{Hyperparameters}.}
			\label{priors_plot}
		\end{figure}
	
		The hyperparameter values we choose for each model can influence their performance, especially when the data is sparse. The hierarchical models try to automatically choose the correct values, by placing a prior over the original hyperparameters. This introduces new hyperparameters, but the models are generally less sensitive to these. 
		
		However, in our experience even the models without hierarchical priors are not very sensitive to this choice, as long as we use fairly weak priors. In particular, we used $\lambda = 0.1$ (GGG, GGGU, GEE, GTT, $\text{GL}^2_1$, GEG), $\eta = \sqrt{10}$ (GLL), and $a = 1, b = 1$ (PGG). The distributions with these hyperparameter values are plotted in Figure \ref{priors_plot}.
		
	\begin{figure*}[t]
		\centering
		\includegraphics[width=0.255\columnwidth]{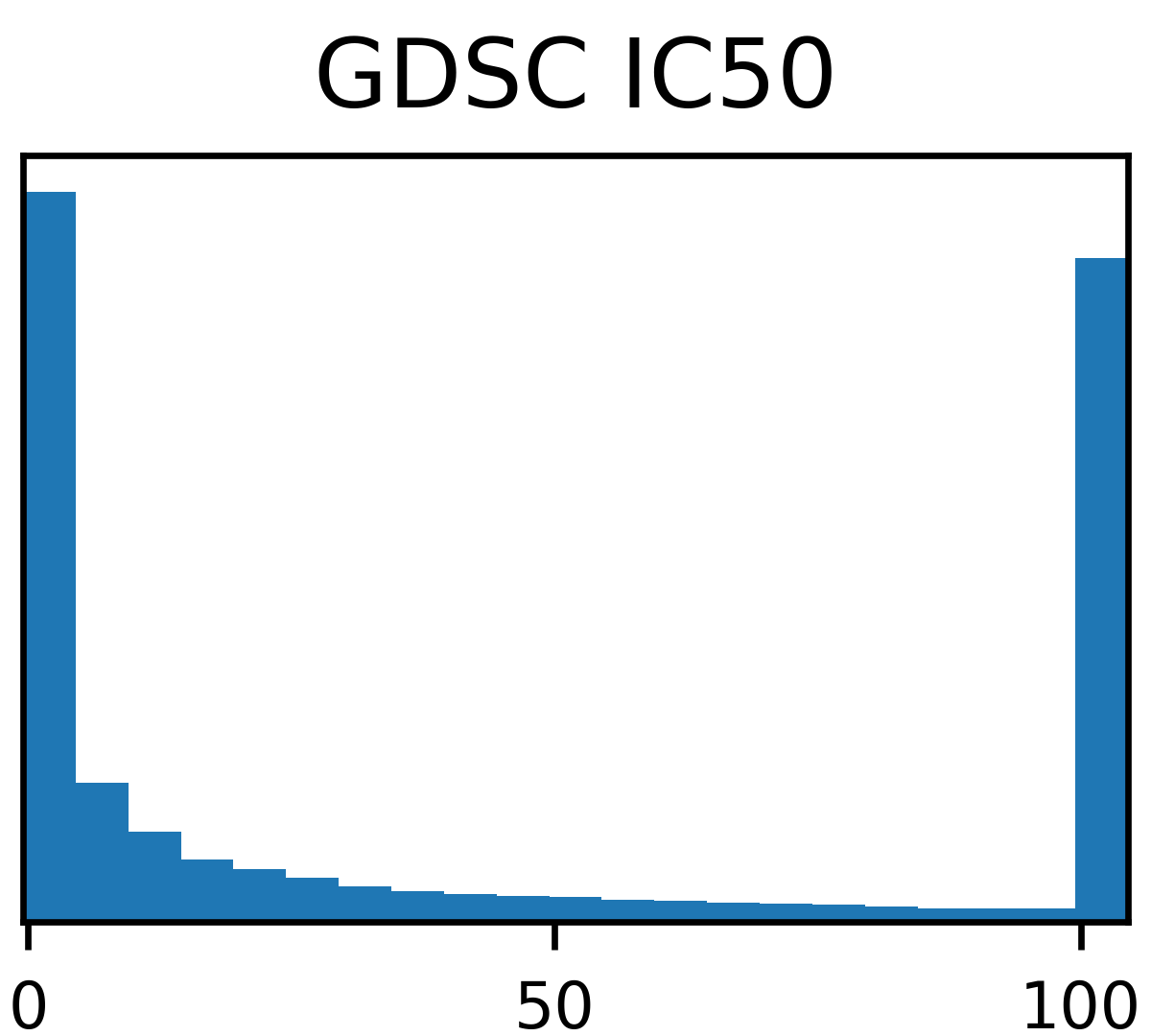}
		\includegraphics[width=0.255\columnwidth]{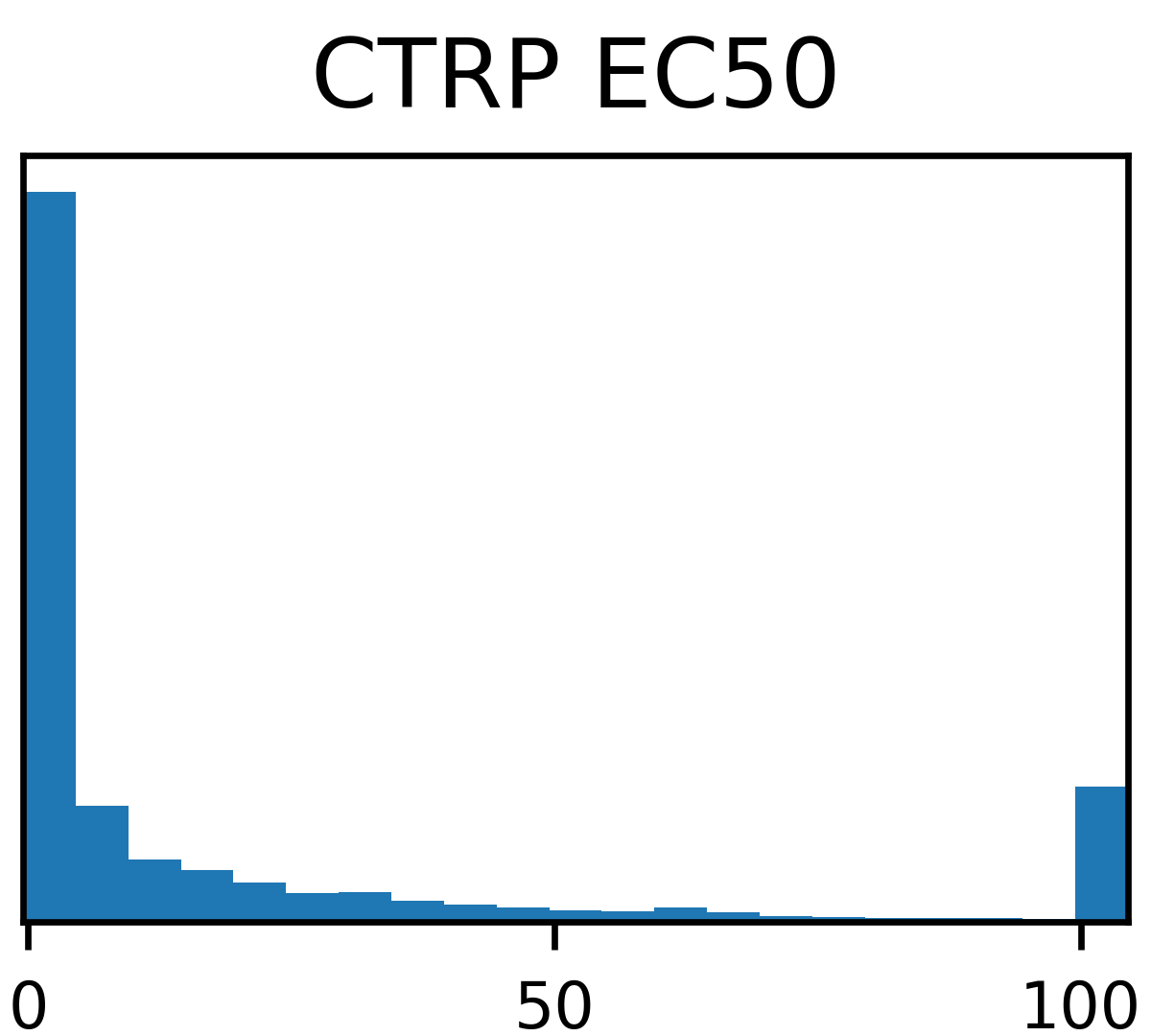}
		\includegraphics[width=0.255\columnwidth]{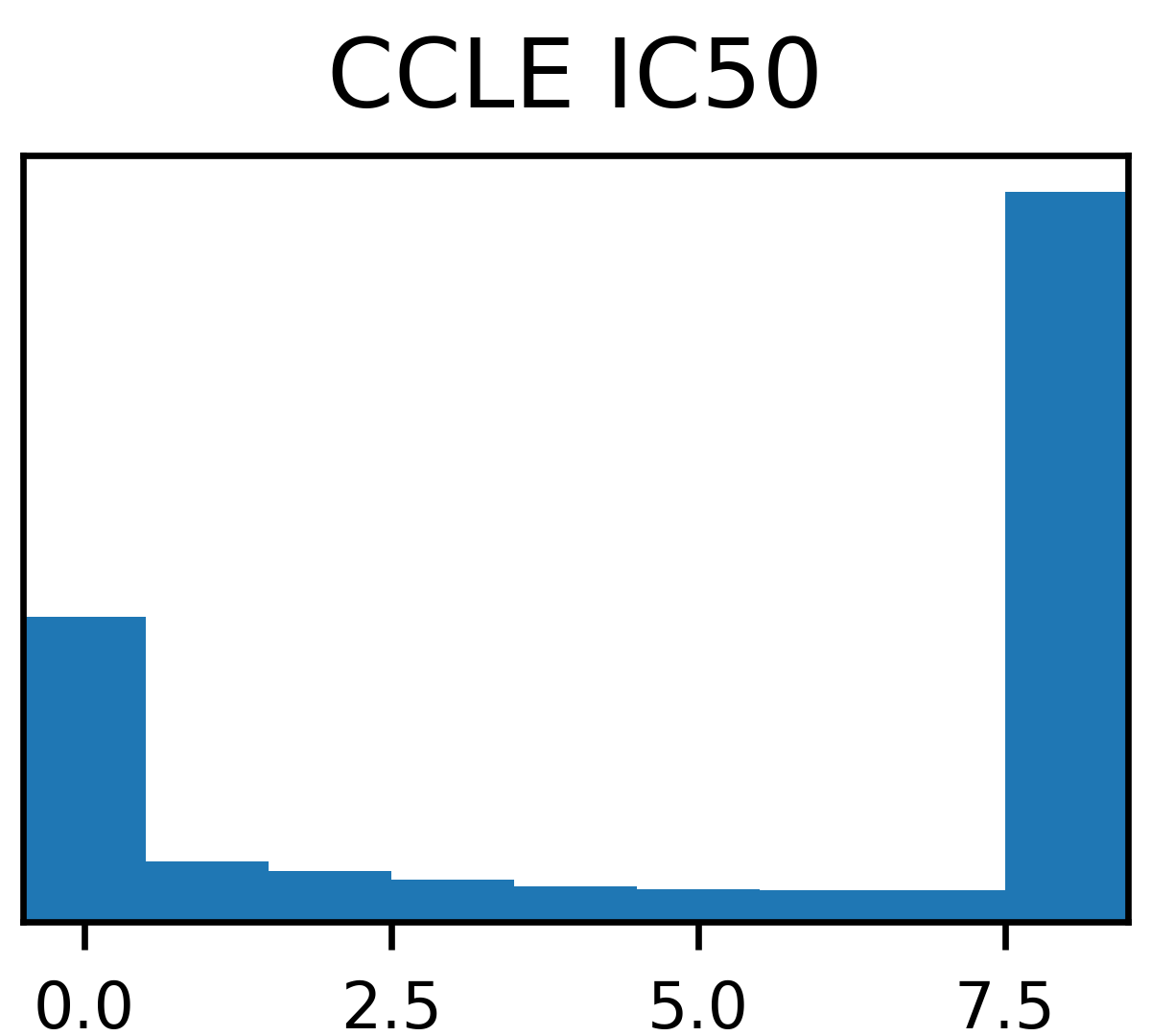}
		\includegraphics[width=0.255\columnwidth]{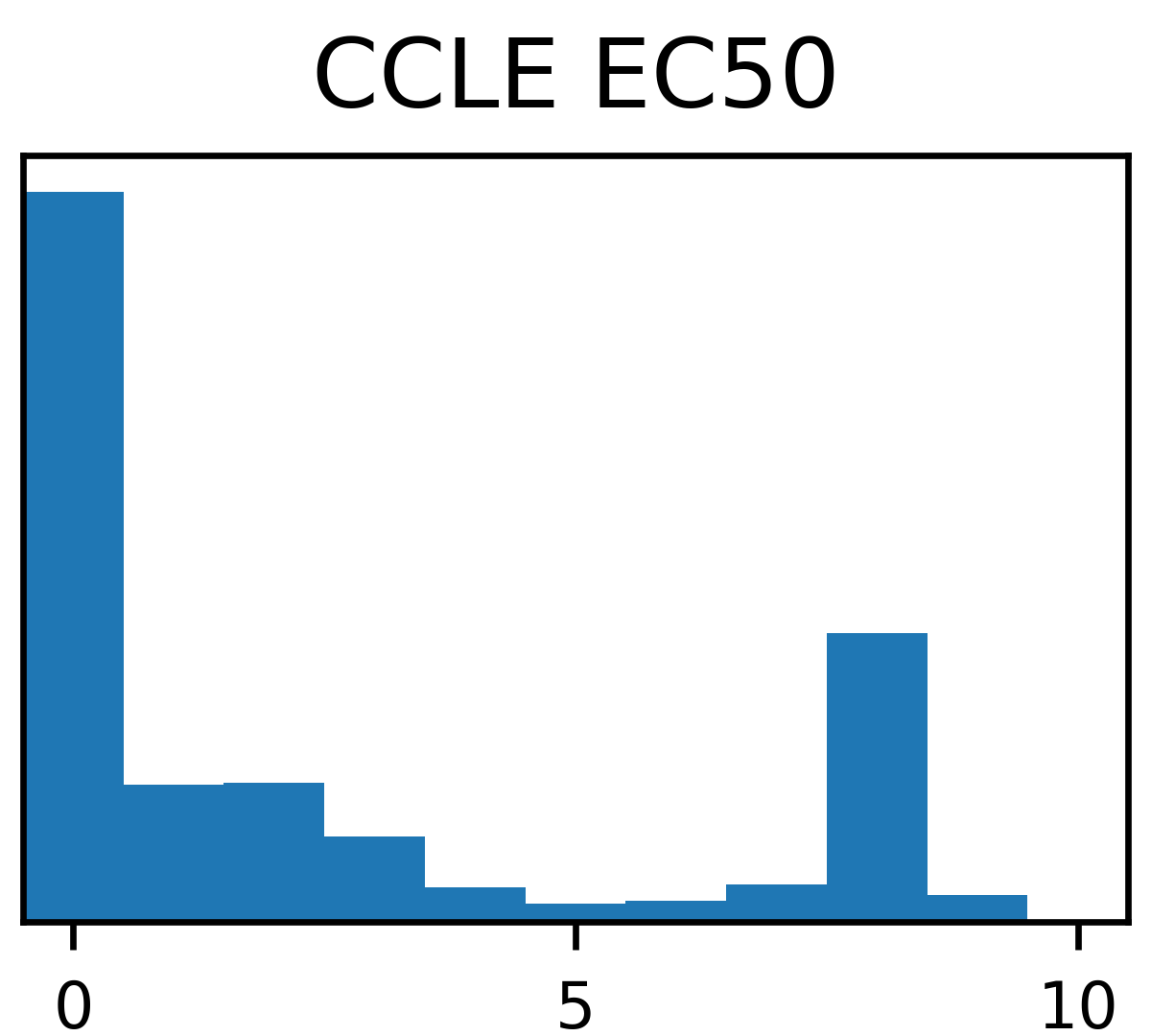}
		\includegraphics[width=0.255\columnwidth]{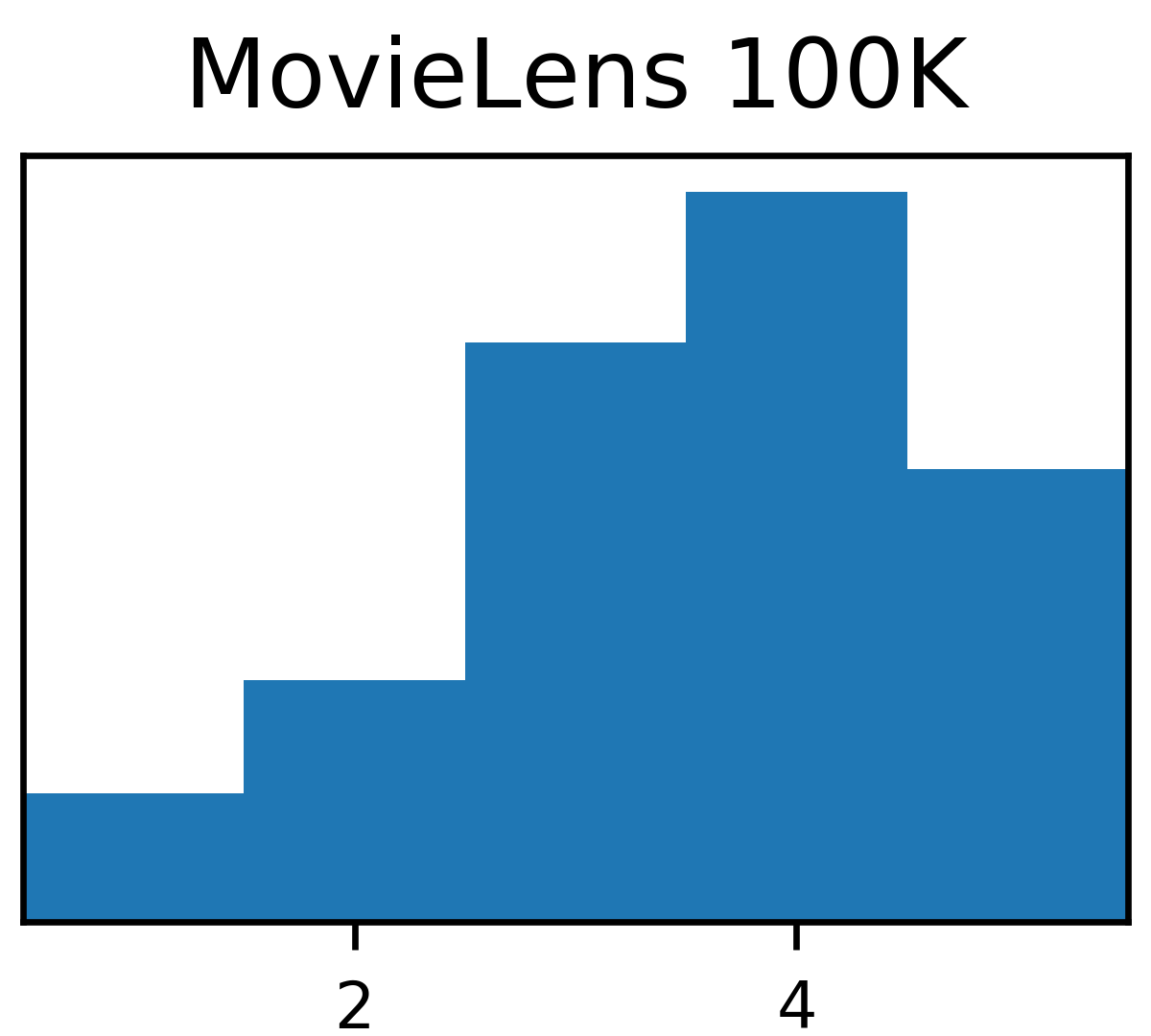}
		\includegraphics[width=0.255\columnwidth]{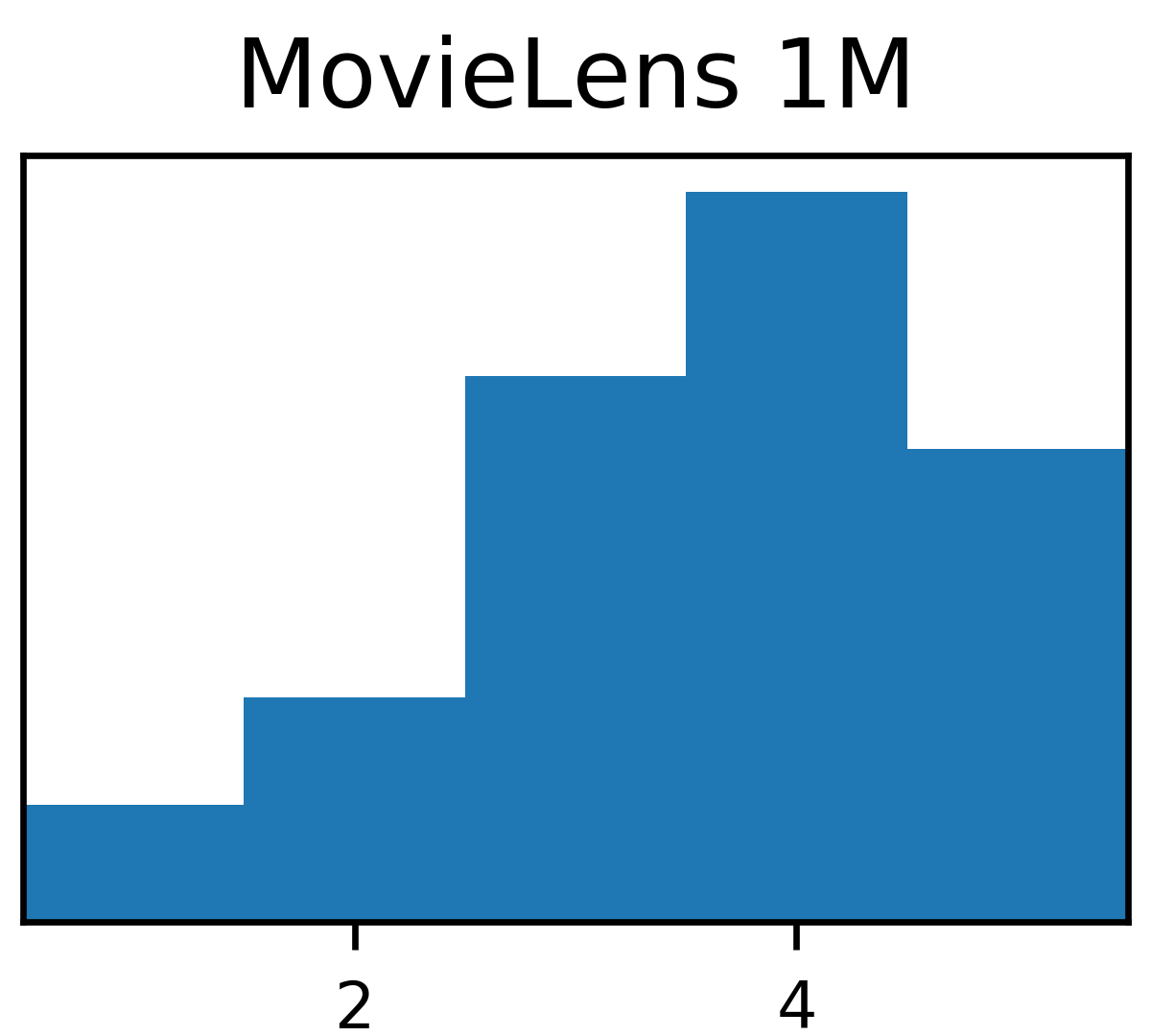}
		\includegraphics[width=0.255\columnwidth]{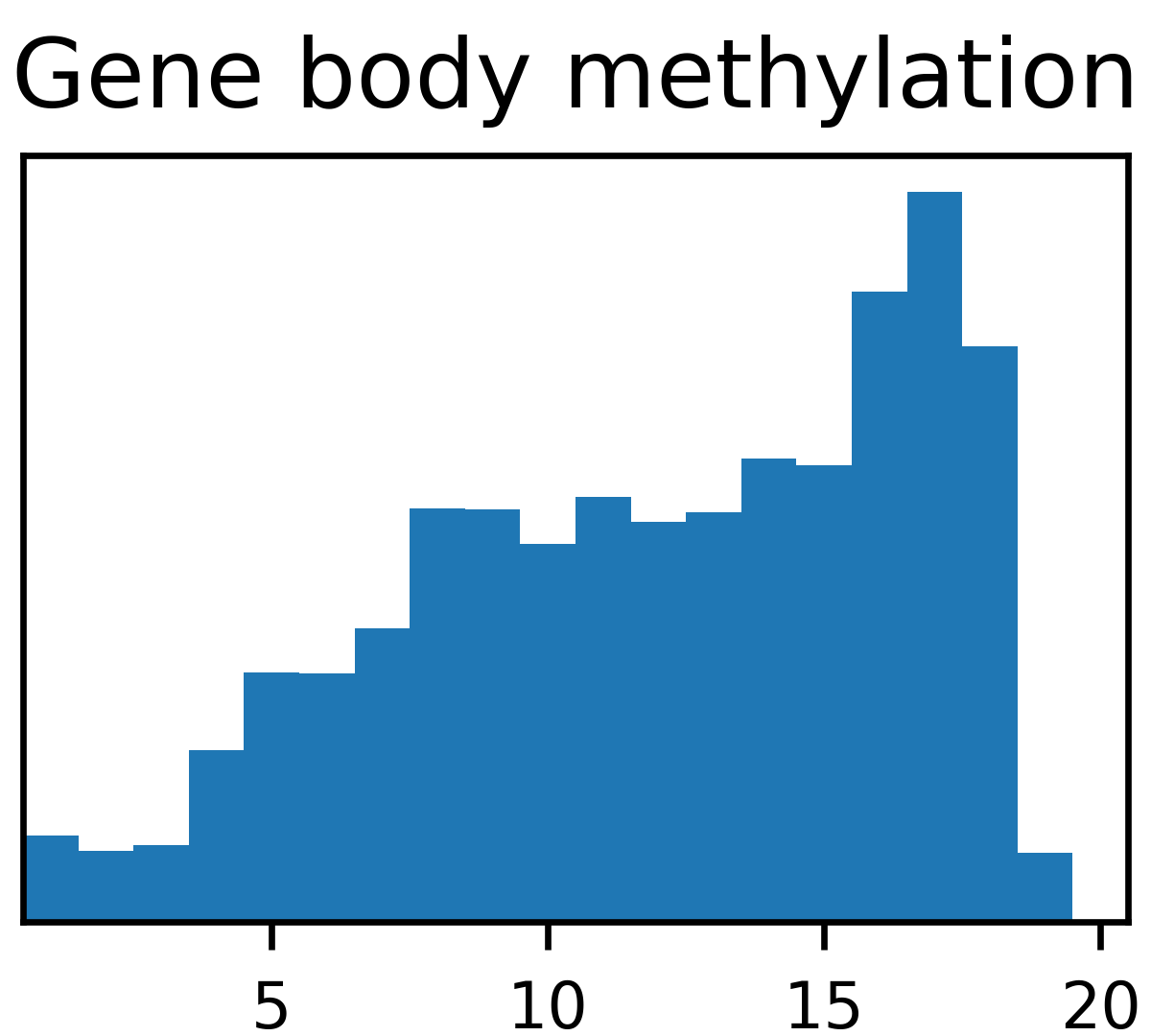}
		\includegraphics[width=0.255\columnwidth]{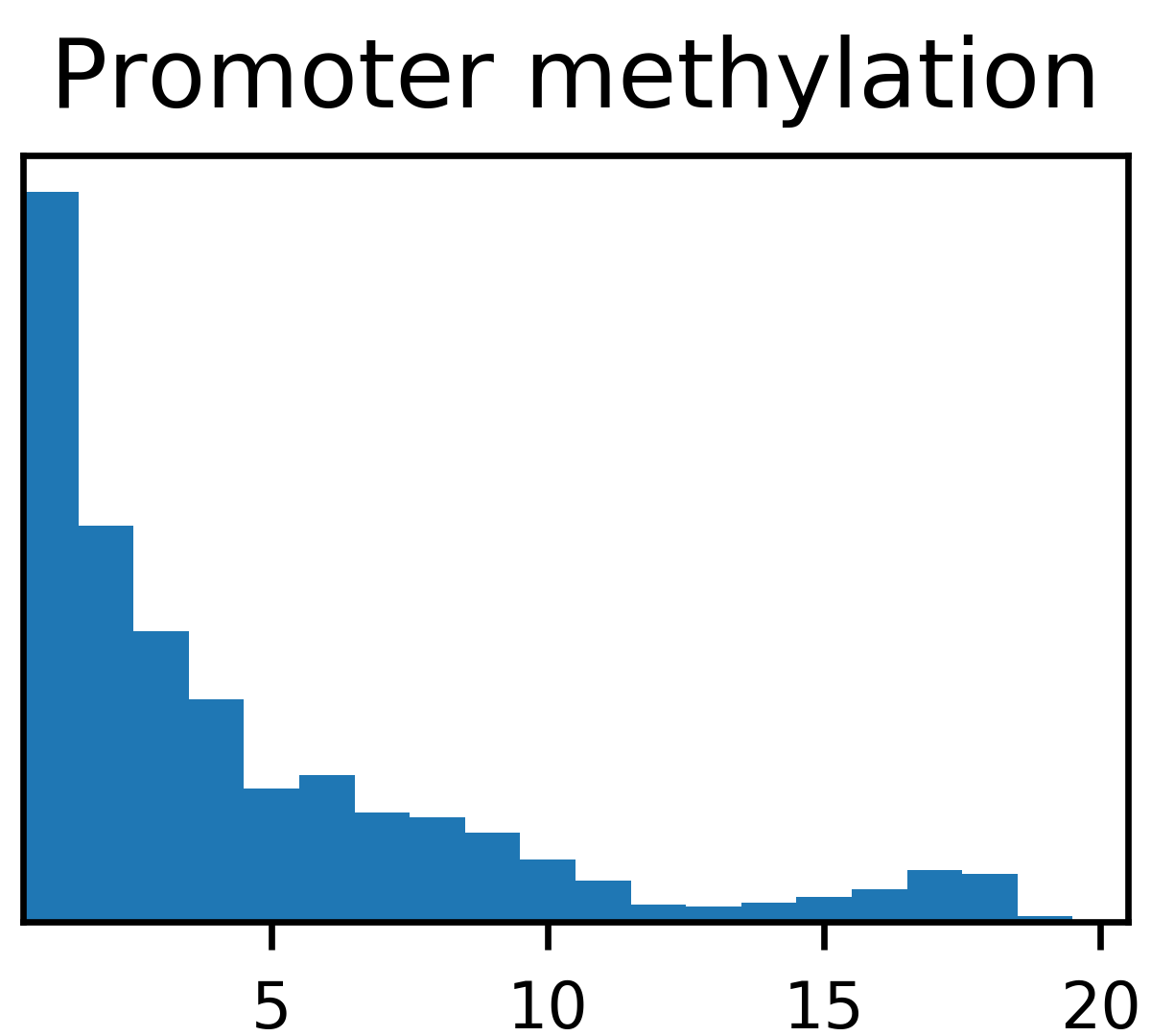}
		\caption{Distributions of the values of the four drug sensitivity, two MovieLens, and two methylation datasets.} 
		\label{dataset_plots}
	\end{figure*}

		For the other models we used:
		$\alpha_{\tau} = \beta_{\tau} = 1 $ (Gaussian likelihood);
		$\alpha_0 = \beta_0 = 1 $ (GGGA, GEEA);
		$\boldsymbol{\mu_0} = \boldsymbol{0}, \beta_0 = 1, \nu_0 = K, \boldsymbol{W_0} = \text{\textbf{I}} $ (GGGW);
		$\mu = \lambda = K $ (GLLI),
		$ \mu_{\mu} = 0, \tau_{\mu} = 0.1, a = b = 1 $ (GTTN),
		$a = a' = b' = 1$ (PGGG).
		
		We did find that the volume prior models (GVG, GVnG) were very sensitive to the hyperparameter choice $\gamma$. The following values were chosen by trying a range on each dataset and choosing the best one: $ \gamma = 10^{\lbrace-30,-20,-10,-10,0,0,0,0\rbrace}$ for \{GDSC,CTRP,CCLE $IC_{50}$,$EC_{50}$,MovieLens 100K,1M,GM,PM\}.

	\subsection{Software}
		Implementations of all models, datasets, and experiments, are available at \url{https://github.com/Anonymous/}.

\section{Datasets}
	We conduct our experiments on a total of eight real-world datasets across three different applications, allowing us to see whether our observations on one dataset or application also hold more generally. We will focus on one or two datasets at a time for more specific experiments. Also note that we make sure all datasets contain only positive integers, so that we can compare all four groups of Bayesian matrix factorisation approaches. 
	
	\begin{table}[t]
		\caption{Overview of the four drug sensitivity, two MovieLens, and two methylation datasets, giving the number of rows (cell lines, users, genes), columns (drugs, movies, patients), and the fraction of entries that are observed.} \label{summary_datasets}
		\centering
		\begin{tabular}{lccc}
			\toprule
			Dataset & Rows & Columns & Fraction obs. \\
			\midrule
			GDSC $IC_{50}$ & 707 & 139 & 0.806 \\
			CTRP $EC_{50}$ & 887 &  545 & 0.801 \\
			CCLE $IC_{50}$ & 504 & 24 & 0.965 \\
			CCLE $EC_{50}$ & 502 & 24 & 0.632 \\
			MovieLens 100K & 943 & 1473 & 0.072 \\
			MovieLens 1M & 6040 & 3503 & 0.047 \\
			Gene body meth. & 160 & 254 & 1.000 \\
			Promoter meth. & 160 & 254 & 1.000 \\
			\bottomrule
		\end{tabular}
	\end{table}
	
	The first comes from bioinformatics, in particular predicting missing values in drug sensitivity datasets, each detailing the effectiveness ($IC_{50}$ or $EC_{50}$ values) of a range of drugs on different cancer and tissue types (cell lines). We consider the Genomics of Drug Sensitivity in Cancer (GDSC v5.0 \cite{Yang2013}, $IC_{50}$), Cancer Therapeutics Response Portal (CTRP v2 \cite{Seashore-Ludlow2015}, $EC_{50}$), and Cancer Cell Line Encyclopedia (CCLE \cite{Barretina2012}, both $IC_{50}$ and $EC_{50}$) datasets. 
	We preprocessed these datasets by: undoing the natural log transform of the GDSC dataset; capping high values to 100 for GDSC and CTRP; and then casting them as integers. We also filtered out rows and columns with only one or two observed datapoints. 
	
	The second application is collaborative filtering, where we are given movie ratings for different users (one to five stars) and we wish to predict the number of stars a user will give to an unseen movie. We use the MovieLens 100K and 1M datasets \cite{Harper2015}, with 100,000 and 1,000,000 ratings respectively.
	
	Finally, another bioinformatics application, this time looking at methylation expression profiles \cite{Koboldt2012}. These datasets give the amount of methylation measured in either the body region of 160 breast cancer driver genes (gene body methylation) or the promoter region (promoter methylation) for 254 different patients. We multiplied all values by twenty and cast them as integers. 
	
	The datasets are summarised in Table \ref{summary_datasets}, and the distribution of values for each dataset is visualised in Figure \ref{dataset_plots}. This shows us that the drug sensitivity datasets tend to be bimodal, whereas the MovieLens and methylation datasets are more normally distributed. We can also see that the MovieLens datasets tend to be large and sparse, whereas the others are well-observed and relatively small.

\section{Experiments} \label{Experiments}
	We conducted experiments to compare the four different groups of approaches. In particular, we measured their convergence speed, cross-validation performance, sparse prediction performance, and model selection effectiveness. We sometimes focus on a selection of the methods for clarity. To make the comparison complete, we also added a popular non-probabilistic nonnegative matrix factorisation model (NMF) \cite{Lee2000} as a baseline.
	The results are discussed in Section \ref{Discussion}.

	\subsection{Convergence}
\begin{figure*}[t]
			\centering
			\begin{minipage}{0.38 \textwidth}
				\centering
				\includegraphics[width=\columnwidth]{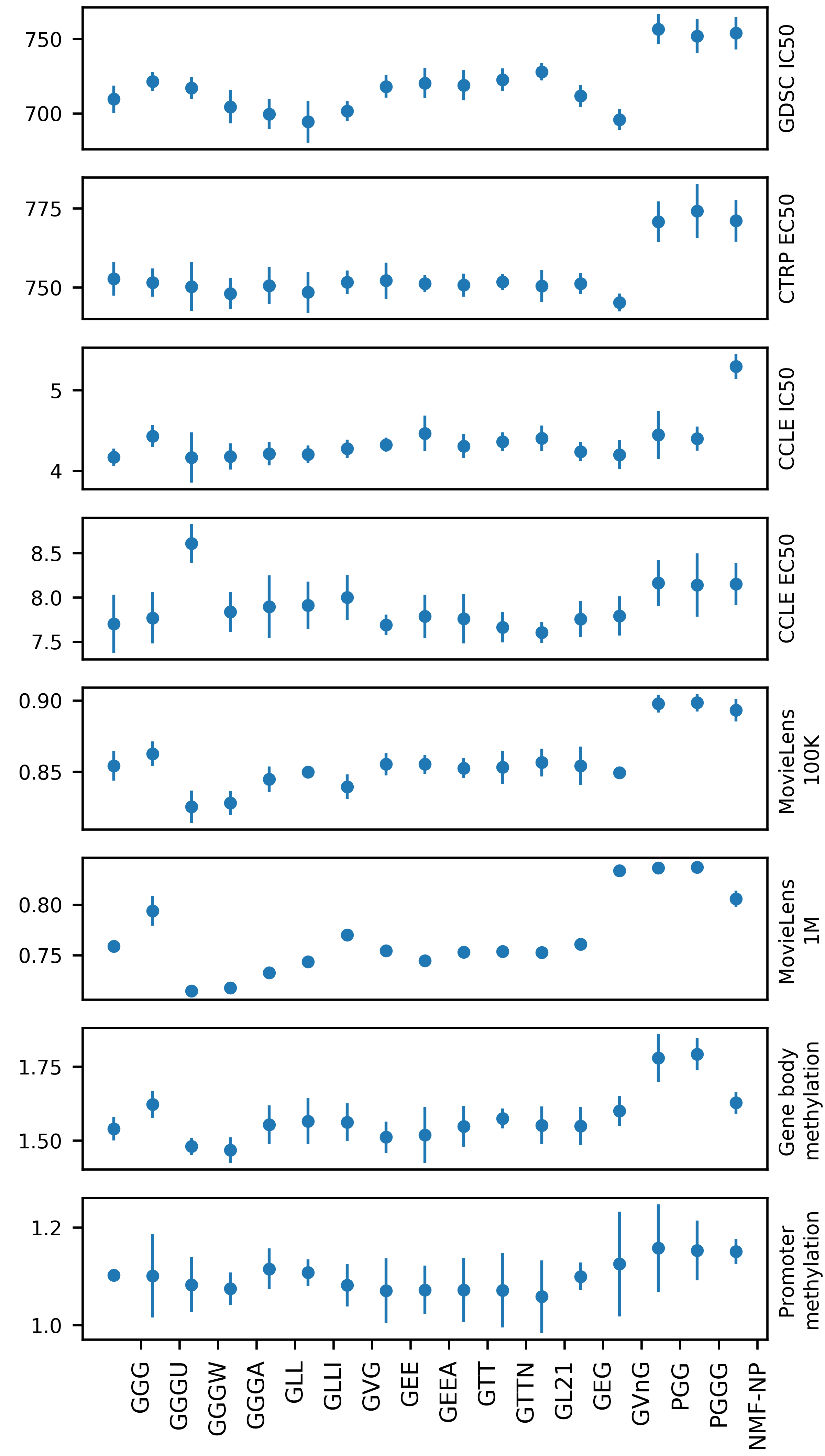}
				\caption{Average mean squared error of 5-fold nested cross-validation for the seventeen methods on the eight datasets. We also plot the standard deviation of errors across the folds.} 
				\label{crossvalidation}
			\end{minipage}
			\hspace{0.02\textwidth}
			\begin{minipage}{0.59 \textwidth}
				\begin{minipage}{\columnwidth}
					\centering
					\includegraphics[width=1\columnwidth]{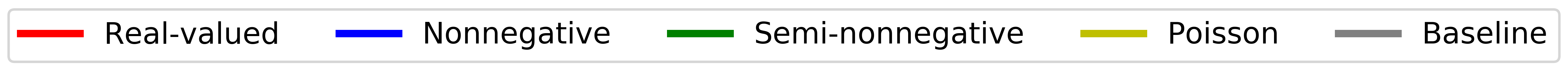}
					
					\vspace{5pt}
					
					\includegraphics[width=0.495\columnwidth]{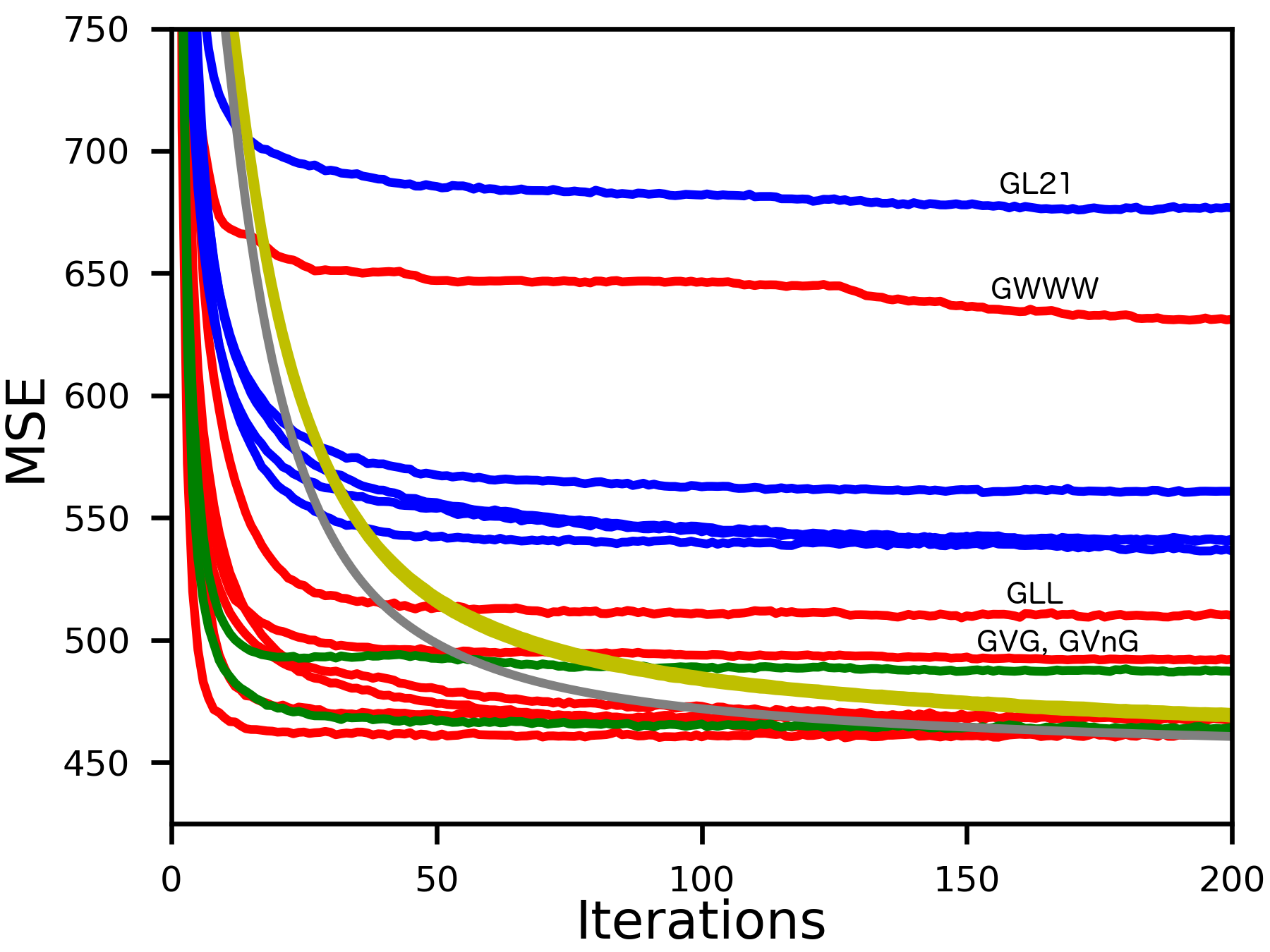}
					\includegraphics[width=0.495\columnwidth]{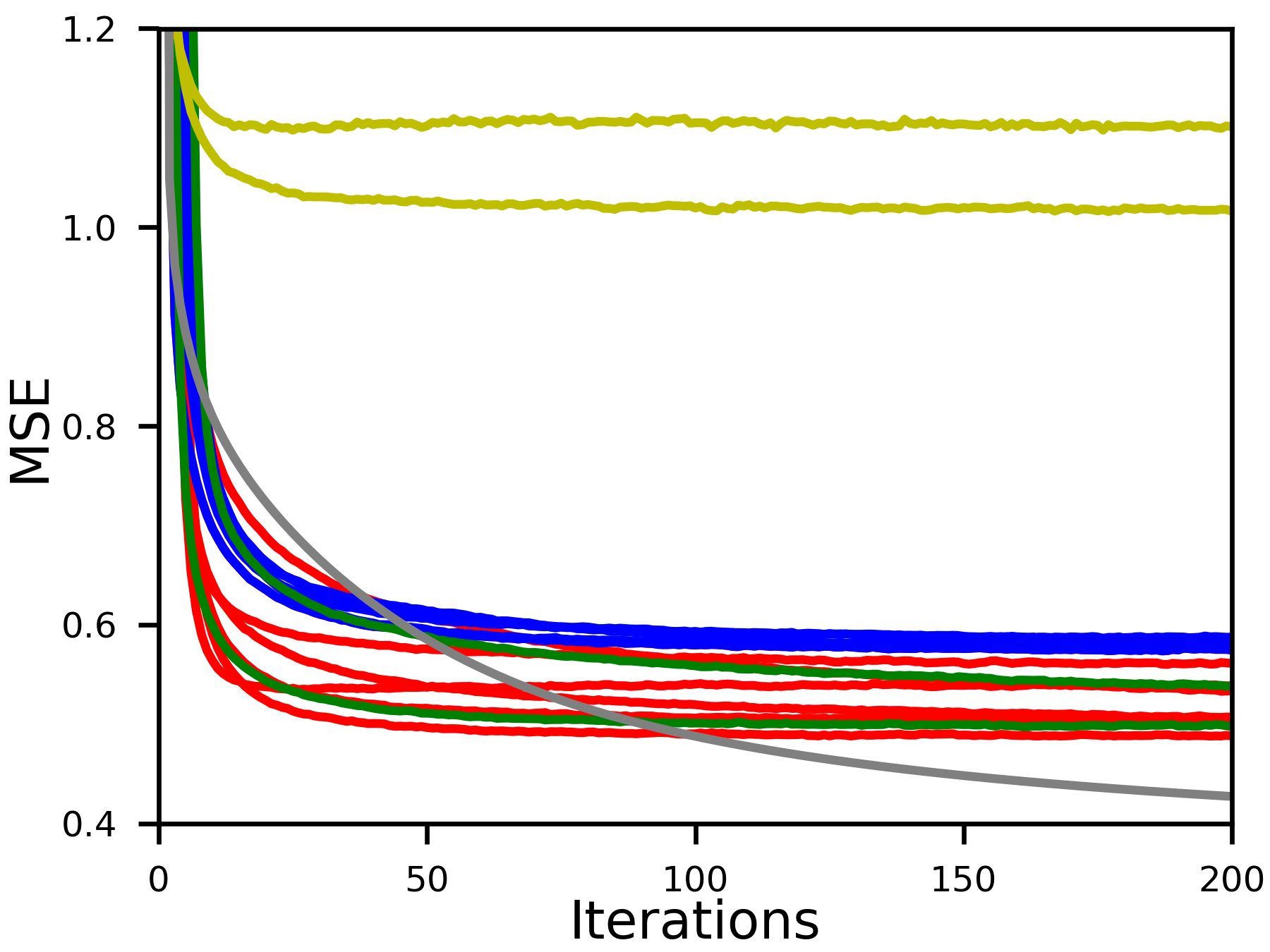}
					\caption{Convergence of the models on the GDSC drug sensitivity (left) and MovieLens 100K (right) datasets, measuring the training data fit (mean square error).} 
					\label{convergences}
				\end{minipage}
			
				\vspace{25pt}
			
				\begin{minipage}{\columnwidth}
					\centering
					\includegraphics[width=\columnwidth]{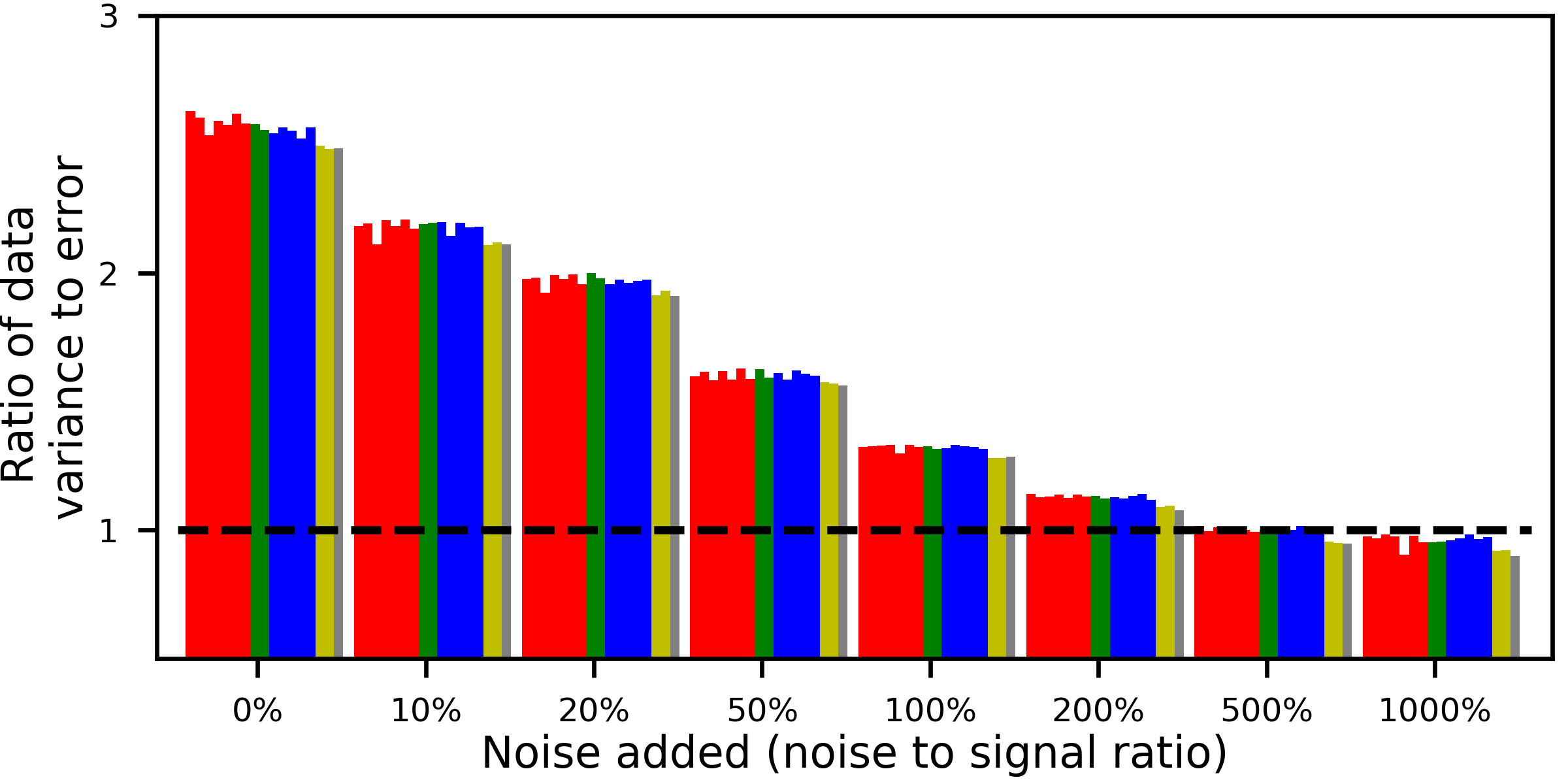}
					\caption{Noise experiment results on the GDSC drug sensitivity dataset. We added different levels of Gaussian noise to the data, and measured the 10-fold cross-validation performance.}
					\label{noise_gdsc}
				\end{minipage}
			\end{minipage}
		\end{figure*}

		Firstly we compared the convergence speed of the models on the GDSC and MovieLens 100K datasets. We ran each model with $K=20$, and measured the average mean squared error on the training data across ten runs. We plotted the results in Figure \ref{convergences}, where each group is plotted as the same colour: red for real-valued, blue for nonnegative, green for semi-nonnegative, yellow for Poisson, and grey for the non-probabilistic baseline.
		Runtime speeds are given in the supplementary materials.
		
	\subsection{Cross-validation} \label{Cross-validation}
		Our first predictive experiment was to measure the 5-fold cross-validation performance on each of the eight datasets. We used the hyperparameter values from Section \ref{Hyperparameters}, and used 5-fold nested cross-validation to choose the dimensionality $K$. The average mean squared error of predictions are given in Figure \ref{crossvalidation} for all eight datasets. The average dimensionality found in nested cross-validation can be found in the supplementary materials.
		
	\subsection{Noise test}
		We then measured the predictive performance when the datasets are very noisy. We added different levels of Gaussian noise to the data, with the noise-to-signal ratio being given by the ratio of the variance of the Gaussian noise we add, to the standard deviation of the generated data. For each noise level we split the datapoints randomly into ten folds, and measured the predictive performance of the models on one held-out set at a time.  We used $K=5$ for all methods. The results for the GDSC drug sensitivity dataset are given in Figure \ref{noise_gdsc}, where we plot the ratio of the variance of the data to the mean squared error of the predictions---higher values are better, and using the row average gives a performance of one. 
		
		\begin{figure*}[t]
			\centering
			\includegraphics[width=0.7\textwidth]{legend_colours.png}
			\vspace{5pt}
			\includegraphics[width=0.9\textwidth]{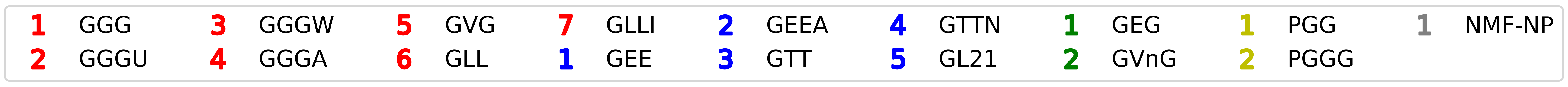}
			\vspace{5pt}
			\begin{minipage}{\textwidth}
				\includegraphics[width=\textwidth]{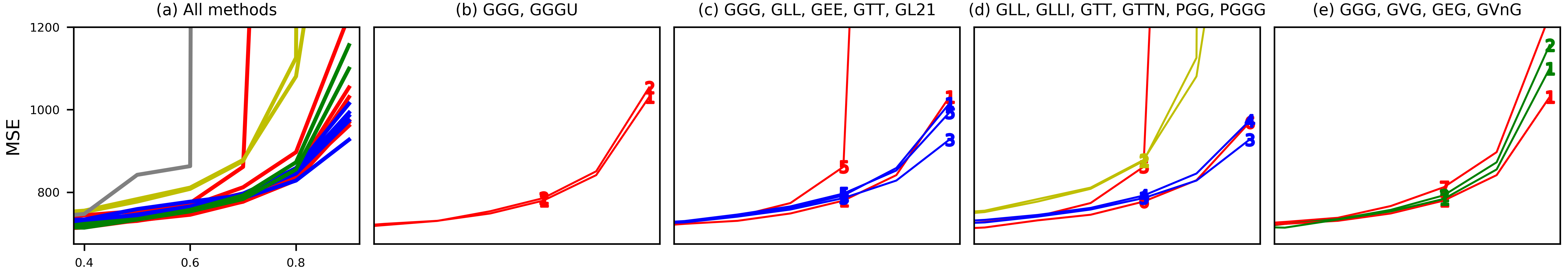}
				
				\includegraphics[width=\textwidth]{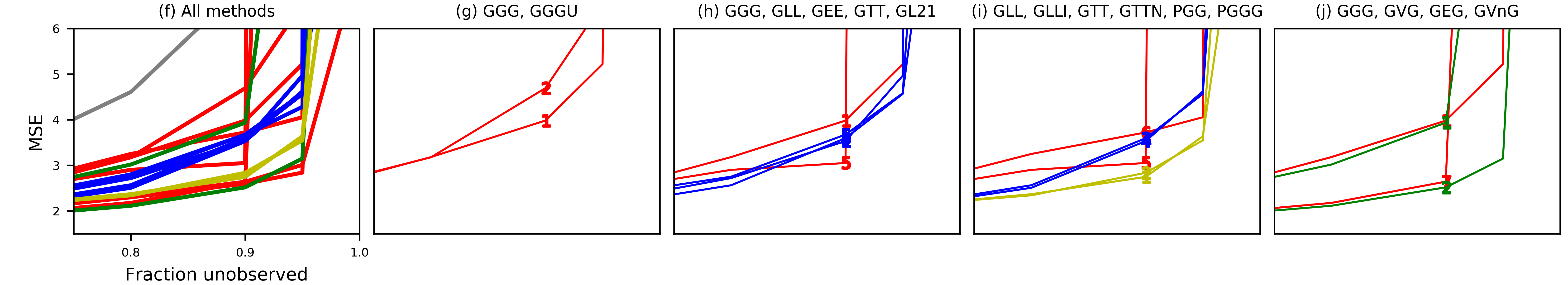}
				\caption{Sparsity experiment results on the GDSC drug sensitivity (top, a-e) and gene body methylation (bottom, f-j) datasets. We measure the predictive performance (mean squared error) on a held-out dataset for different fractions of unobserved data.}
				\label{sparsity_gm_gdsc}
			\end{minipage}
			\begin{minipage}{\textwidth}
				\includegraphics[width=\textwidth]{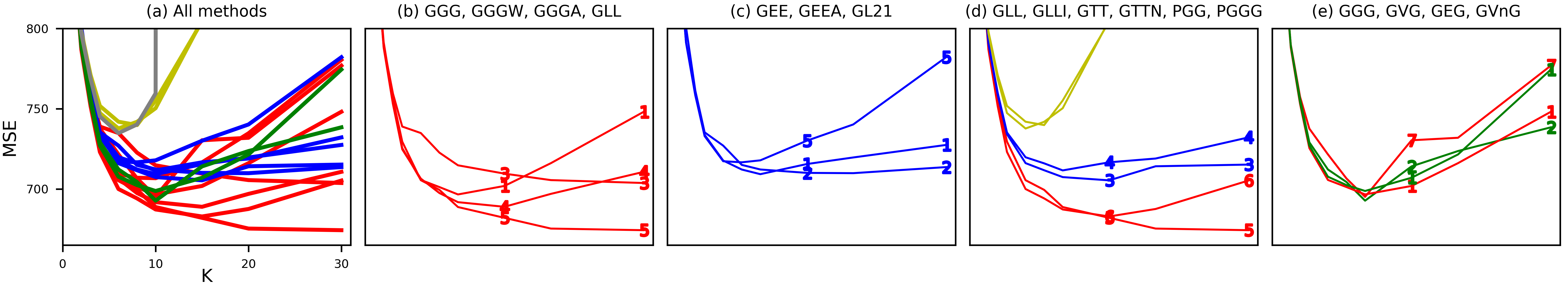}
				\caption{Model selection experiment results on the GDSC drug sensitivity dataset. We measure the predictive performance (mean squared error) on a held-out dataset for different dimensionalities $K$.}
				\label{model_selection_gdsc}
			\end{minipage}
		\end{figure*}
	
	\subsection{Sparse predictions}
		Next we measured the predictive performances when the sparsity of the data increases. For different fractions of unobserved data, we randomly split the data based on that fraction, trained the model on the observed data, and measured the performance on the held-out test data. We used $K=5$ for all models.
		The average mean squared error of ten repeats is given in Figure \ref{sparsity_gm_gdsc}, showing the performances on both the methylation GM and GDSC drug sensitivity datasets.
	
	\subsection{Model selection}
		We also measured the robustness of the models to overfitting if the dimensionality $K$ is high. As a result, most models will fit very well to the training data, but give poor predictions on the test data. Here, we vary the dimensionality $K$ for each of the models on the GDSC drug sensitivity dataset, randomly taking out $10\%$ as test data, and repeating ten times.
		The results are given in Figure \ref{model_selection_gdsc}---in the supplementary materials we look at two more datasets.

\section{Discussion} \label{Discussion}
	From the results shown in the previous section, we were able to draw the following conclusions. 
	
	Observation 1: Poisson likelihood methods perform poorly compared to the Gaussian likelihood---they overfit quickly (Figures \ref{model_selection_gdsc}a), give worse predictive performances in cross-validation (Figure \ref{crossvalidation}) and under noisy conditions (Figure \ref{noise_gdsc}), presumably because they cannot converge as deep as the other methods (Figure \ref{convergences}). At high sparsity levels they can start to perform better (Figure \ref{sparsity_gm_gdsc}d). Some papers \cite{Gopalan2015} claim that Poisson models offer better predictions, but for small and well-observed datasets we found the opposite to be true. 
	
	Observation 2: Nonnegative models are more constrained than the real-valued ones, causing them to converge less deep (Figure \ref{convergences}), and to be less likely to overfit to high sparsity levels (\ref{sparsity_gm_gdsc}c, \ref{sparsity_gm_gdsc}h) than the standard GGG model. However, the right hierarchical prior for a real-valued model (such as Wishart) can bridge this gap.
	
	Observation 3: There is no difference in performance between real-valued and semi-nonnegative matrix factorisation, as shown in the model selection and sparsity experiments (Figures \ref{sparsity_gm_gdsc}e, \ref{sparsity_gm_gdsc}j, and \ref{model_selection_gdsc}e): the performance for GGG and GEG, as well as GVG and GVnG, are nearly identical.
		
	Observation 4: There is no difference in predictive performance between univariate and multivariate posteriors (GGG, GGGU), as shown in Figures \ref{sparsity_gm_gdsc}b and \ref{sparsity_gm_gdsc}g.
		
	Observation 5: The automatic relevance determination and Wishart hierarchical priors are effective ways of preventing overfitting, as shown in Figures \ref{model_selection_gdsc}b and \ref{model_selection_gdsc}c: the GGGA, GGGW, and GEEA models keep the line down as $K$ increase, whereas the GGG and GEE models start overfitting more. This has been shown before for nonnegative models \cite{Brouwer2017b} but the effect is even stronger for the real-valued ones. 
		
	Overvation 6: Similarly, the Laplace priors are good at reducing overfitting as the dimensionality grows (Figure \ref{model_selection_gdsc}b), without requiring additional hierarchical priors.
		
	Observation 7: Some other hierarchical priors do not make a difference, such as with GLLI, GTTN, PGGG---Figures \ref{sparsity_gm_gdsc}d, \ref{sparsity_gm_gdsc}i, and \ref{model_selection_gdsc}d show little difference in performance. They can help us automatically choose the hyperparameters, but in our experience the models are not very sensitive to this choice anyways. 
	
	Although these observations are specific to the applications and dataset sizes studied, we believe that general insights can be drawn from them about the behaviour of the four different groups of Bayesian matrix factorisation models. The behaviour of Poisson models is especially interesting, because they are often claimed to be better than Gaussian models for large datasets, but for smaller ones this does not hold. We hope that these insights will assist future researchers in their model design.

\bibliography{bibliography}

\end{document}


\title{Prior and Likelihood Choices for Bayesian Matrix Factorisation on Small Datasets \\ \vspace{10pt} Supplementary Materials}
	\author{}
	\date{}
	\maketitle{}
	
	\noindent \large{Thirty-Second AAAI Conference on Artificial Intelligence (2018).} \\
	
	\setcounter{secnumdepth}{3}
	\setcounter{tocdepth}{2}
	\tableofcontents
	\clearpage
		
		
	\section{Gibbs sampling algorithms}
		In this section we give the Gibbs sampling posteriors for the sixteen Bayesian matrix factorisation models studied in the paper. 
		
		The idea of Gibbs sampling is as follows. We wish to sample values from the posterior distribution $p(\btheta|D)$, but we cannot sample these directly. Instead, we could draw 2values from the conditional posterior $ p(\theta_i | \btheta_{-i}, D ) $, which is the distribution over the parameter $ \theta_i $ (such as $U_{ik}$) given the current values of the other parameters $ \btheta_{-i} $, and the observed data $ D $. If we sample new values in turn for each parameter $ \theta_i $ from $ p(\theta_i | \btheta_{-i}, D ) $, we will eventually converge to draws from the true posterior $ p(\btheta|D) $, which can be used to approximate it. 
		In this paper we focus on models where we have so-called \textit{model conjugacy}, allowing us to sample from the conditional posteriors.
		
		We give the Gibbs sampling posterior distributions for $\U$, which can be derived using Bayes' theorem. For example, the derivations for $U_{ik}$ in the GEE model can be found below. Expressions for $\V$ are symmetrical, and hence omitted. 
		\begin{alignat*}{1}
			p(U_{ik}|\tau,\U_{-ik},\V,\boldsymbol \lambda) 
			&\propto p(\R|\tau,\U,\V) \times p(U_{ik}|\lambda_k) \\
			&\propto \prod_{j \in \Omega^1_i} \mathcal{N} (R_{ij} | \U_i \cdot \V_j, \tau^{-1} ) \times \mathcal{E} ( U_{ik} | \lambda_k) \\
			&\propto \exp \left\{ - \frac{\tau}{2} \sum_{j \in \Omega^1_i} (R_{ij} - \U_i \V_j)^2 \right\} \times \exp \left\{ - \lambda_k U_{ik} \right\} \times u(x) \\
			&\propto \exp \left\{ - \frac{U_{ik}^2}{2} \left[ \displaystyle \tau \sum_{j \in \Omega^1_i} V_{jk}^2 \right] \right. \\
			& \left. \hspace{46pt} + U_{ik} \left[ - \lambda_k + \tau \sum_{j \in \Omega^1_i} \diffexclk V_{jk} \right] \right\} \times u(x) \\
			&\propto \exp \left\{ - \frac{\tauUik}{2} ( U_{ik} - \muUik )^2 \right\} \times u(x) \\
			&\propto \mathcal{TN} ( U_{ik} | \muUik, \tauUik ).
		\end{alignat*}
	
		\subsection{Real-valued matrix factorisation}
		The first group of methods use a Gaussian likelihood for noise, and place real-valued prior distributions over $\U$ and $\V$, typically Gaussian as well. We assume each value in $\R$ comes from the product of $\U$ and $\V$, with Gaussian noise added,
		\begin{alignat*}{2}
			&R_{ij} \sim \mathcal{N} (R_{ij} | \Ui \Vj, \tau^{-1} )	 \quad\quad		&&\tau \sim \mathcal{G} (\tau | \alpha_{\tau}, \beta_{\tau} ) 
		\end{alignat*}
		where $ \mathcal{N} (x|\mu,\tau) = \tau^{\frac{1}{2}} (2\pi)^{-\frac{1}{2}} \exp \left\{ -\frac{\tau}{2} (x - \mu)^2 \right\} $ is the density of the Gaussian distribution, with precision $ \tau $. $\Ui, \Vj$ denote the $i$th and $j$th rows of $\U$ and $\V$. We place a further Gamma prior over $\tau$, with $ \mathcal{G} (\tau | \alpha_{\tau}, \beta_{\tau} ) = \frac{{\beta_{\tau}}^{\alpha_{\tau}}}{\Gamma(\alpha_{\tau})} x^{\alpha_{\tau} -1} e^{- \beta_{\tau} x} $, where $ \Gamma(x) = \int_{0}^{\infty} x^{t-1} e^{-x} dt $ is the gamma function. 
		
		In the Gibbs sampling algorithm, the posterior for the noise parameter is
		\begin{alignat*}{1}
			\tau \sim \mathcal{G} (\tau | \alpha^*_{\tau}, \beta^*_{\tau} )		\quad\quad 		\alpha^*_{\tau} = \alpha_{\tau} + \frac{\vert \Omega \vert}{2} 		\quad\quad  		\beta^*_{\tau} = \beta_{\tau} + \frac{1}{2} \sumk \diff^2.
		\end{alignat*}
		
		\subsubsection{All Gaussian model (GGG)}
		We place Gaussian independent priors over the entries in $\U,\V$, with hyperparameter $\lambda$,
		\begin{alignat*}{1}
			\Ui \sim \mathcal{N} ( \Ui | \boldsymbol 0, \lambda^{-1} \text{\textbf I} )		\quad\quad	\Vj \sim \mathcal{N} ( \Vj | \boldsymbol 0, \lambda^{-1} \text{\textbf I} )
		\end{alignat*}
		where $ \mathcal{N} (\boldsymbol x|\boldsymbol \mu,\boldsymbol \Sigma) = \vert \boldsymbol \Sigma \vert^{-\frac{1}{2}} (2\pi)^{-\frac{K}{2}} \exp \left\{ -\frac{1}{2} (\boldsymbol x - \boldsymbol \mu)^T \boldsymbol \Sigma^{-1} (\boldsymbol x - \boldsymbol \mu) \right\} $ is the density of a $K$-dimensional multivariate Gaussian distribution, and \text{\textbf I} is the identity matrix.
		The conditional posterior distributions we obtain in the Gibbs sampling algorithm are also multivariate Gaussians. The parameter values are given below, with $ \Omega_i = \left\{ j \text{ $ \vert $ } (i,j) \in \Omega \right\} $ and  $ \Omega_j = \left\{ i \text{ $ \vert $ } (i,j) \in \Omega \right\} $. 
		\begin{alignat*}{1}
			&\U_i \sim \mathcal{N} ( \Ui | \muUi, \SigmaUi )		
			\quad\quad	\muUi = \SigmaUi \cdot \left[ \tau \sumOmegai R_{ij} \Vj \right] 		
			\quad\quad	 \SigmaUi = \left[ \lambda \text{\textbf I} + \tau \sumOmegai \left( \Vj \otimes \Vj \right) \right]^{-1}
		\end{alignat*}
		
		\subsubsection{All Gaussian model with univariate posterior (GGGU)}
		It is also possible to have a univariate posterior for the Gibbs sampler,
		\begin{alignat*}{1}
			&U_{ik} \sim \mathcal{N} (U_{ik} | \mu^U_{ik}, (\tau^U_{ik})^{-1} )
			\quad\quad \mu^U_{ik} = \frac{1}{\tau^U_{ik}} \Bigg[ \tau \sumOmegai \diffexclk V_{jk} \Bigg] 
			\quad\quad \tau^U_{ik} = \lambda + \tau \sumOmegai V_{jk}^2.
		\end{alignat*}
		
		\subsubsection{All Gaussian model with ARD hierarchical prior (GGGA)}
		For the automatic relevance determination (ARD) prior we replace the $\lambda$ hyperparameter by a factor-specific variable $\lambda_k$, which has a further Gamma prior. 
		\begin{alignat*}{1}
			\Ui \sim \mathcal{N} ( \Ui | \boldsymbol 0, \text{diag}(\boldsymbol \lambda^{-1}) )		\quad\quad	\Vj \sim \mathcal{N} ( \Vj | \boldsymbol 0, \text{diag}(\boldsymbol \lambda^{-1}) ) 	\quad\quad 		\lambda_k \sim \mathcal{G} (\lambda_k | \alpha_0, \beta_0 )
		\end{alignat*}
		where $\text{diag}(\boldsymbol \lambda^{-1})$ is a diagonal matrix with entries $\lambda_1^{-1}, .., \lambda_K^{-1}$ on the diagonal.
		The Gibbs sampling posteriors for $\Ui$ and $\Vj$ are largely unchanged, replacing $\lambda \text{\textbf I}$ in the expressions for $\SigmaUi, \SigmaVj$ by $\text{diag}(\lambda_1,..,\lambda_K)$. The posterior for $\lambda_k$ is
		\begin{alignat*}{1}
			\lambda_k \sim \mathcal{G} (\lambda_k | \alpha^*_0, \beta^*_0 )		\quad\quad 		\alpha^*_0 = \alpha_0 + \frac{I}{2} + \frac{J}{2} 		\quad\quad  		\beta^*_0 = \beta_0 + \frac{1}{2} \sum_{i=1}^I U_{ik}^2 + \frac{1}{2} \sum_{j=1}^J V_{jk}^2.
		\end{alignat*}
		
		\subsubsection{All Gaussian model with Wishart hierarchical prior (GGGW)}
		Instead of assuming independence of each entry in $\U, \V$, we now assume each row of $\U$ comes from a multivariate Gaussian with row mean $\boldsymbol{\mu_U}$ and covariance $\boldsymbol{\Sigma_U}$, and similarly for $\V$. We place a further Normal-Inverse Wishart prior over these parameters,
		\begin{alignat*}{2}
			&\Ui \sim \mathcal{N} ( \Ui | \boldsymbol{\mu_U}, \boldsymbol{\Sigma_U})		\quad\quad	&&\boldsymbol{\mu_U}, \boldsymbol{\Sigma_U} \sim \mathcal{NIW} ( \boldsymbol{\mu_U}, \boldsymbol{\Sigma_U} | \boldsymbol{\mu_0}, \beta_0, \nu_0, \boldsymbol{W_0} ) \\
			&\Vj \sim \mathcal{N} ( \Vj | \boldsymbol{\mu_V}, \boldsymbol{\Sigma_V})		\quad\quad 		&&\boldsymbol{\mu_V}, \boldsymbol{\Sigma_V} \sim \mathcal{NIW} ( \boldsymbol{\mu_V}, \boldsymbol{\Sigma_V} | \boldsymbol{\mu_0}, \beta_0, \nu_0, \boldsymbol{W_0} )
		\end{alignat*}
		where $ \mathcal{NIW} ( \boldsymbol{\mu_U}, \boldsymbol{\Sigma_U} | \boldsymbol{\mu_0}, \beta_0, \nu_0, \boldsymbol{W_0} ) = \mathcal{N}(\boldsymbol \mu | \boldsymbol{\mu_0}, \frac{1}{\beta_0} \text{\textbf I} ) \mathcal{W}^{-1} ( \boldsymbol \Sigma | \nu_0, \boldsymbol{W_0} ) $ is the density of a normal-inverse Wishart distribution, and $\mathcal{W}^{-1} ( \boldsymbol \Sigma | \nu_0, \boldsymbol{W_0} )$ is the inverse Wishart distribution. 
		
		For the Gibbs sampling algorithm we obtain the posteriors $\Ui \sim \mathcal{N} ( \Ui | \muUi, \SigmaUi )$ and $ \boldsymbol{\mu_U}, \boldsymbol{\Sigma_U} \sim \mathcal{NIW} ( \boldsymbol{\mu_U}, \boldsymbol{\Sigma_U} | \boldsymbol{\mu_0^*}, \beta_0^*, \nu_0^*, \boldsymbol{W_0^*} ) $, with
		\begin{alignat*}{1}
			&\muUi = \SigmaUi \cdot \Big[ \boldsymbol{\Sigma_U^{-1}}  \boldsymbol{\mu_U} + \tau \sumOmegai R_{ij} \Vj \Big] 
			\quad\quad \SigmaUi = \Big[ \boldsymbol{\Sigma_U^{-1}} + \tau \sumOmegai \left( \Vj \otimes \Vj \right) \Big]^{-1} \\
			&\beta_0^* = \beta_0 + I  
			\quad\quad \nu_0^* = \nu_0 + I
			\quad\quad \boldsymbol{\mu_0^*} = \frac{\beta_0 \boldsymbol{\mu_0} + I \boldsymbol{\bar{U}}}{\beta_0 + I} 
			\quad\quad\quad \boldsymbol{\bar{U}} = \frac{1}{I} \sum_{i=1}^I \Ui \\
			&\boldsymbol{W_0^*} = \boldsymbol{W_0} + I \boldsymbol{\bar{S}} + \frac{\beta_0 I}{\beta_0 + I} ( \boldsymbol{\mu_0} - \boldsymbol{\bar{U}} ) \otimes ( \boldsymbol{\mu_0} - \boldsymbol{\bar{U}} )
			\quad\quad\quad\quad \boldsymbol{\bar{S}} = \frac{1}{I} \sum_{i=1}^I ( \Ui \otimes \Ui ).
		\end{alignat*}
		
		\subsubsection{Gaussian likelihood with Laplace priors (GLL)}
		An alternative to the Gaussian prior is to use the Laplace distribution, which has a much more pointy distribution than Gaussian around $x=0$. This leads to more sparse solutions, as more factors are set to low values. The priors are
		\begin{alignat*}{1}
			U_{ik} \sim \mathcal{L} ( U_{ik} | 0, \eta ) 
			\quad\quad V_{jk} \sim \mathcal{L} ( V_{jk} | 0, \eta )
		\end{alignat*}
		To simplify inference we introduce a new variable $\lambdaUik$ for each $U_{ik}$, with prior $\lambdaUik \sim \mathcal{E} (\lambdaUik | \eta)$. The idea behind this is that we can rewrite a Laplace distribution as
		\begin{alignat*}{1}
			\mathcal{L} ( x | \mu, \rho ) = \int_{\epsilon=0}^{\infty} \mathcal{N}(x|\mu,\epsilon) \mathcal{E}(\epsilon|\frac{\rho}{2}) d\epsilon
		\end{alignat*}
		This leads to the following Gibbs sampling posteriors:
		\begin{alignat*}{2}
			&\U_i \sim \mathcal{N} ( \Ui | \muUi, \SigmaUi )		
			\quad\quad	&& \muUi = \SigmaUi \cdot \left[ \tau \sumOmegai R_{ij} \Vj \right] \\
			& && \SigmaUi = \text{diag}((\boldsymbol{\lambda^U_i})^{-1}) + \left[ \tau \sumOmegai \left( \Vj \otimes \Vj \right) \right]^{-1} \\
			& \frac{1}{\lambdaUik} \sim \mathcal{IG}(x|\mu^U_{ik}, \lambda^U_{ik})
			\quad\quad && \mu^U_{ik} = \frac{\sqrt{\eta}}{|U_{ik}|}
			\quad\quad\quad\quad \lambda^U_{ik} = \eta.
		\end{alignat*}
		
		\subsubsection{Gaussian likelihood with Laplace and hierarchical inverse Gaussian priors (GLLI)}
		We can place a further hierarchical prior over the $\eta$ parameters, 
		\begin{alignat*}{1}
			\eta^U_{ik} \sim \mathcal{IG}(\mu, \lambda)  
			\quad\quad \eta^V_{jk} \sim \mathcal{IG}(\mu, \lambda).
		\end{alignat*}
		The paper that introduced this prior (Jing, Wang and Yang 2015) placed a Generalised Inverse Gaussian $\mathcal{GIG}(\gamma, a, b)$ prior over the $\eta$ parameters, but then used $\gamma = -\frac{1}{2}$, which reduces the prior to the Inverse Gaussian above with $\mu=\sqrt{b/a}, \lambda=b$ (or $a=1/\mu, b=\lambda$).
		
		The posteriors for $\U, \V$ are identical, and for $\lambdaUik$ we only replace $\eta$ with $\eta^U_{ik}$. We obtain another Inverse Gaussian posterior for the $\eta^U_{ik}$ parameters,
		\begin{alignat*}{1}
			\eta^U_{ik} \sim \mathcal{IG} (\eta^U_{ik} | \mu^{\eta}_{ik}, \lambda^{\eta}_{ik} )
			\quad\quad \mu^U_{ik} = \sqrt{\frac{\lambdaUik + a}{b}} = \sqrt{\frac{\lambdaUik + 1/\mu}{\lambda}}
			\quad\quad \lambda^U_{ik} = \lambdaUik + b = 1/\mu.
		\end{alignat*}
		
		\subsubsection{Gaussian likelihood with volume prior (GVG)}
		The prior over $\V$ in this model is Gaussian, as in the GGG model, but we now use the volume prior (VP) for the $\U$ matrix, with density $p(\U) \propto \exp \lbrace - \gamma \det (\U^T \U) \rbrace $. This model leads to the posterior 
		\begin{alignat*}{2}
			&U_{ik} \sim \mathcal{N} (U_{ik} | \mu^U_{ik}, (\tau^U_{ik})^{-1} )
			\quad\quad &&\mu^U_{ik} = \frac{1}{\tau^U_{ik}} \Bigg[ \gamma \boldsymbol{U_{i\tilde{k}}} \boldsymbol{A_{\tilde{k}\tilde{k}}} ( \boldsymbol{U_{\tilde{i}\tilde{k}}^T} \boldsymbol{U_{\tilde{i}k}} ) + \tau \sumOmegai \diffexclk V_{jk} \Bigg] \\
			& &&\tau^U_{ik} = \tau \sumOmegai V_{jk}^2 + \gamma ( D_{\tilde{k}\tilde{k}} - \boldsymbol{U_{i\tilde{k}}} \boldsymbol{A_{\tilde{k}\tilde{k}}} \boldsymbol{U_{i\tilde{k}}^T} ).
		\end{alignat*}
		In the above, vector $ \boldsymbol{U_{i\tilde{k}}} $ is the $i$th row of $\U$ excluding column $k$; vector $ \boldsymbol{U_{\tilde{i}k}} $ is the $k$th column of $\U$ excluding row $i$; matrix $ \boldsymbol{U_{\tilde{i}\tilde{k}}} $ is $\U$ excluding row $i$ and column $k$; matrix $ \boldsymbol{U_{\cdot \tilde{k}}} $ is $\U$ excluding column $k$; $D_{\tilde{k}\tilde{k}} = \det \lbrace \boldsymbol{U_{\cdot \tilde{k}}^T} \boldsymbol{U_{\cdot \tilde{k}}} \rbrace $; and matrix $\boldsymbol{A_{\tilde{k}\tilde{k}}} = \det \lbrace \boldsymbol{U_{\cdot \tilde{k}}^T} \boldsymbol{U_{\cdot \tilde{k}}} \rbrace $ is the matrix adjugate.
		
		\subsection{Nonnegative matrix factorisation}
		Nonnegative matrix factorisation models use the same Gaussian noise model as the real-valued ones, but placing nonnegative prior distributions over entries in $\U$ and $\V$. 
		
		\subsubsection{Gaussian likelihood with exponential priors (GEE)}
		This model places independent Exponential priors over the entries in $\U,\V$,
		\begin{alignat*}{1}
			&U_{ik} \sim \mathcal{E} ( U_{ik} | \lambda )		\quad\quad	V_{jk} \sim \mathcal{E} ( V_{jk} | \lambda ).
		\end{alignat*}
		The product of a Gaussian and Exponential distribution leads to a truncated normal posterior,
		\begin{alignat*}{2}
			& U_{ik} \sim \mathcal{TN} (U_{ik} | \mu^U_{ik}, \tau^U_{ik} )
		\quad\quad &&\mu^U_{ik} = \frac{1}{\tau^U_{ik}} \Bigg[ - \lambda + \tau \sumOmegai \diffexclk V_{jk} \Bigg]
		\quad\quad \tau^U_{ik} = \tau \sumOmegai V_{jk}^2.
		\end{alignat*}
		
		\subsubsection{Gaussian likelihood with exponential prior and ARD (GEEA)}
		Similar to the GGGA model, we can extend GEE with the ARD prior,
		\begin{alignat*}{1}
			&U_{ik} \sim \mathcal{E} ( U_{ik} | \lambda_k )		\quad\quad	V_{jk} \sim \mathcal{E} ( V_{jk} | \lambda_k ) 	\quad\quad 		\lambda_k \sim \mathcal{G} (\lambda_k | \alpha_0, \beta_0 )
		\end{alignat*}
		The posteriors for $\U$ and $\V$ are the same as in the GEE model, but replacing $\lambda$ by $\lambda_k$. The posteriors for $\lambda_k$ become
		\begin{alignat*}{1}
			\lambda_k \sim \mathcal{G} (\lambda_k | \alpha^*_0, \beta^*_0 )		\quad\quad 		\alpha^*_0 = \alpha_0 + I + J 		\quad\quad  		\beta^*_0 = \beta_0 + \sum_{i=1}^I U_{ik} + \sum_{j=1}^J V_{jk}.
		\end{alignat*}

		\subsubsection{Gaussian likelihood with truncated normal priors (GTT)}
		We can also use the truncated normal distribution directly as the priors for $\U$ and $\V$, 
		\begin{alignat*}{1}
			&U_{ik} \sim \mathcal{TN} ( U_{ik} | \mu_U, \tau_U )		\quad\quad	V_{jk} \sim \mathcal{TN} ( V_{jk} | \mu_V, \tau_V ) 
		\end{alignat*}
		This again gives a truncated normal posterior, but with slightly different values.
		\begin{alignat*}{2}
			& U_{ik} \sim \mathcal{TN} (U_{ik} | \mu^U_{ik}, (\tau^U_{ik})^{-1} )
			\quad\quad &&\mu^U_{ik} = \frac{1}{\tau^U_{ik}} \Bigg[ \mu_U \tau_U + \tau \sumOmegai \diffexclk V_{jk} \Bigg] \\
			& &&\tau^U_{ik} = \tau_U + \tau \sumOmegai V_{jk}^2.
		\end{alignat*}
		An alternative to the truncated normal distribution is the so-called half normal. If random variable $y$ has density $\mathcal{N} (y | 0, \sigma^2 )$, and random variable $x = |y|$, then $x$ follows a half normal distribution with density $\mathcal{HN}(x|\sigma) = \frac{\sqrt{2}}{\sigma \sqrt{\pi}} \exp \lbrace - \frac{x^2}{2\sigma^2} \rbrace u(x) $. Note however that when $\mu = 0$ in the truncated normal distribution, then these distributions are equivalent, with $\tau = \frac{1}{\sigma^2} $, so the GTT model is more general.
		
		\subsubsection{Gaussian likelihood with truncated normal and hierarchical priors (GTTN)}
		We can place a further prior over the parameters of the truncated normal distributions,
		\begin{alignat*}{1}
			&U_{ik} \sim \mathcal{TN} ( U_{ik} | \mu^U_{ik}, \tau^U_{ik} )		\quad\quad	V_{jk} \sim \mathcal{TN} ( V_{jk} | \mu^V_{jk}, \tau^V_{jk} )  \\
			&p(\mu^U_{ik}, \tau^U_{ik} | \mu_{\mu}, \tau{\mu}, a, b) \propto \frac{1}{\sqrt{\tau^U_{ik}}} \left( 1 - \Phi ( - \mu^U_{ik} \sqrt{\tau^U_{ik}} ) \right) \mathcal{N} (\mu^U_{ik} | \mu_{\mu}, \tau_{\mu}^{-1} ). \mathcal{G} (\tau^U_{ik} | a, b)
		\end{alignat*}
		The density for $\mu^V_{jk}, \tau^V_{jk}$ is identical. Note that this is not the same as the product of a Normal and Gamma distribution. It is not easy to sample from this prior, but it can be used as a hierarchical prior. The posteriors for $U_{ik}$ remain the same (replacing $\mu_U$ and $\tau_U$ by $\mu^U_{ik}$ and $\tau^U_{ik}$), and for $\mu^U_{ik}$ and $ \tau^U_{ik}$ we obtain posteriors
		\begin{alignat*}{3}
			& \mu^U_{ik} \sim \mathcal{N} (\mu^U_{ik} | m_{\mu}, t_{\mu}^{-1} )
			\quad\quad && m_{\mu} = \frac{1}{t_{\mu}} \left[ \tau^U_{ik} U_{ik} + \mu_{\mu} \tau_{\mu} \right] 
			\quad\quad && t_{\mu} = \tau^U_{ik} + \tau_{\mu} \\
			& \tau^U_{ik} \sim \mathcal{G} (\tau^U_{ik} | a^*, b^* )
			\quad\quad && a^* = a + \frac{1}{2} 
			\quad\quad && b^* = b + \frac{(U_{ik} - \mu^U_{ik})^2}{2}.
		\end{alignat*}
		
		\subsubsection{Gaussian likelihood with $\boldsymbol{\text{L}^2_1}$ norm priors (G$\boldsymbol{\text{L}^2_1}$)}
		We can also use a prior inspired by the $\text{L}^2_1$ norm for both $\U$ and $\V$, giving prior densities
		\begin{alignat*}{1}
			& p ( \U ) \propto \left\{
			\begin{array}{ll}
			\displaystyle \exp \lbrace -\frac{\lambda}{2} \sum_i \left( \sum_k U_{ik} \right)^2 \rbrace   & \mbox{if } U_{ik} \geq 0 \mbox{ for all $i, k$} \\
			0 & \mbox{if any } U_{ik} < 0
			\end{array}
			\right. \\
			& p ( \V ) \propto \left\{
			\begin{array}{ll}
			\displaystyle \exp \lbrace -\frac{\lambda}{2} \sum_j \left( \sum_k V_{jk} \right)^2   & \mbox{if } V_{jk} \geq 0 \mbox{ for all $j, k$} \\
			0 & \mbox{if any } V_{jk} < 0
			\end{array}
			\right.
		\end{alignat*}
		The nonnegativity constraint is used to address the fact that the $\text{L}^2_1$ norm used the absolute value of entries in $\U, \V$, which makes inference impossible unless we constrain them to be nonnegative (in which case the values are automatically absolute).
		
		The posteriors are similar to the GEE and GTT models, but now adding a term that depends on the other entries in the $i$th (or $j$th) row of $\U$,
		\begin{alignat*}{2}
			& U_{ik} \sim \mathcal{TN} (U_{ik} | \mu^U_{ik}, \tau^U_{ik} )
			\quad\quad &&\mu^U_{ik} = \frac{1}{\tau^U_{ik}} \Bigg[ - \lambda \sumexclk U_{ik'} + \tau \sumOmegai \diffexclk V_{jk} \Bigg] \\
			& && \tau^U_{ik} = \lambda + \tau \sumOmegai V_{jk}^2.
		\end{alignat*}
		
		\subsection{Semi-nonnegative matrix factorisation}
		In the nonnegative matrix factorisation models we placed nonnegative priors over both $\U$ and $\V$.
		Instead, we could constrain only one to be nonnegative. In the Bayesian setting this is done by placing a real-valued prior over one matrix, and a nonnegative prior over the other. The major advantage is that we can handle real-valued datasets, while still enforcing some nonnegativity. 
		
		\subsubsection{Gaussian likelihood with nonnegative volume prior (GVnG)}
		The volume prior discussed earlier can also be formulated to be nonnegative. In particular, the probability distribution over $\U$ is
		\begin{equation*}
			p ( \U ) \propto \left\{
			\begin{array}{ll}
			\displaystyle \exp \lbrace - \gamma \det (\U^T \U) \rbrace   & \mbox{if } U_{ik} \geq 0 \mbox{ for all $i, k$} \\
			0 & \mbox{if any } U_{ik} < 0
			\end{array}
			\right.
		\end{equation*}
		The posterior parameters are the same as for the GVG model, but drawing new values from a truncated normal, rather than normal. For $\V$ we again use a Gaussian.
		
		\subsubsection{Gaussian likelihood with exponential and Gaussian priors (GEG)}
			For this model we use an exponential prior for entries in $\U$, and a Gaussian for $\V$. The posteriors are given in the GEE and GGG model sections.

		\subsection{Poisson matrix factorisation}
		The final category of matrix factorisation models do not use a Gaussian likelihood, instead opting for a Poisson one. This only works for nonnegative count data, with $ \R \in \mathbb{N}^{I \times J} $. We again assume each value in $\R$ comes from the product of $\U$ and $\V$, $ R_{ij} \sim \mathcal{P} (R_{ij} | \Ui \Vj )$, where $ \mathcal{P} (x|\lambda) = \frac{\lambda^x \exp \lbrace - \lambda \rbrace}{x!} $ is the density of a Poisson distribution. 
		
		\subsubsection{Poisson likelihood with Gamma priors (PGG)}
		The standard Poisson matrix factorisation model uses independent Gamma priors over the entries in $\U$ and $\V$. To make inference simpler, we also introduce random variables $Z_{ijk}$ such that $R_{ij} = \sumk Z_{ijk}$, each effectively accounting for the contribution of factor $k$ to $R_{ij}$. We use the following distributions and priors: 
		\begin{alignat*}{1}
			Z_{ijk} \sim \mathcal{P} (Z_{ijk} | U_{ik} V_{jk} ) 		\quad\quad 			U_{ik} \sim \mathcal{G} ( U_{ik} | a, b )		\quad\quad	V_{jk} \sim \mathcal{G} ( V_{jk} | a, b ).
		\end{alignat*}
		Note that we can do this because the sum of Poisson distributed random variables (like $Z_{ijk}$) is again Poisson distributed, with rate $\lambda $ equal to the sum of rates of the $Z_{ijk}$, giving us the original Poisson likelihood for $R_{ij}$. The above is also equivalent to saying that $\boldsymbol{Z_{ij}} \sim \text{Mult} (\boldsymbol{Z_{ij}} | n, \boldsymbol p) $ with $ n = R_{ij} $ and $ \boldsymbol p = (\frac{U_{i1} V_{j1}}{\Ui \Vj}, .., \frac{U_{iK} V_{jK}}{\Ui \Vj}) $, where $\boldsymbol{Z_{ij}}$ is a vector containing $Z_{ij1}, .., Z_{ijK}$, and $\text{Mult}(\boldsymbol x | n, \boldsymbol p) = \frac{n!}{x_1! .. x_K!} p_1^{x_1} .. p_K^{x_K} $ is a K-dimensional multinomial distribution. 
		
		Using the above trick, the posteriors are
		\begin{alignat*}{3}
			& \boldsymbol{Z_{ij}} \sim \text{Mult} (\boldsymbol{Z_{ij}} | n, \boldsymbol{p} )
			\quad\quad && n = R_{ij}
			\quad\quad && \boldsymbol{p} = (\frac{U_{i1} V_{j1}}{\Ui \Vj}, .., \frac{U_{iK} V_{jK}}{\Ui \Vj}) \\
			& U_{ik} \sim \mathcal{P} (U_{ik} | a^*_{ik}, b^*_{ik} )
			\quad\quad && a^*_{ik} = a + \sumOmegai Z_{ijk}
			\quad\quad &&b^*_{ik} = b + \sumOmegai V_{jk}.
		\end{alignat*}
		
		\subsubsection{Poisson likelihood with Gamma and hierarchical Gamma priors  (PGGG)}
		We can extend the standard Poisson matrix factorisation model with hierarchical priors. The priors are
		\begin{alignat*}{1}
			&U_{ik} \sim \mathcal{G} ( U_{ik} | a, h^U_i )		\quad\quad			V_{jk} \sim \mathcal{G} ( V_{jk} | a, h^V_j )		\quad\quad			h^U_i \sim \mathcal{G} (a', \frac{a'}{b'}) 		\quad\quad		h^V_j \sim \mathcal{G} (a', \frac{a'}{b'})
		\end{alignat*}
		The posteriors for $U_{ik}$ are identical to the PGG model, except replacing $b$ with $h^U_i$ in the expression for $b^*_{ik}$. For the hierarchical part we obtain the following posteriors:
		\begin{alignat*}{3}
		&h^U_i \sim \mathcal{G} (h^U_i | a^*_i, b^*_i )   		
		\quad\quad && a^*_i = a' + K a 		
		\quad\quad && b^*_i = \frac{a'}{b'} + \sumk U_{ik}.
		\end{alignat*}
	
	
	\clearpage
	\section{Computational complexity}
		The different matrix factorisation models have different time complexities for computing the parameter values and sample new values for $\U$, $\V$, and any other random variables. The space complexity for all models is $\mathcal{O}( I K + J K ) $ per iteration, with an additional $ K \vert \Omega \vert $ term for the Poisson models (for the $Z_{ijk}$).
		
		The time complexities per iteration for the multivariate Gaussian posterior models (GGG, GGGA, GGGW, GLL, GLLI) is $ \mathcal{O}( (I+J)K^3 + IJK^2 ) $. However, these row draws and parameter value computations can all be done in parallel. 
		The univariate posterior models (GGGU, GEE, GEEA, GTT, GTTN, $\text{GL}^2_1$, GEG) have complexity $ \mathcal{O}( I J K^2 ) $, but the parameters can be computed efficiently per column. The volume prior models (GVG, GVnG) have the highest complexity, with $ \mathcal{O}( I^2 J K^2 ) $. Finally, the Poisson models are $ \mathcal{O}( I J K ) $, but this hides a big constant that effectively makes it the slowest model for low values of $K$.
			
		
	\section{Runtime speed}
		The average runtime (in seconds) per iteration is given in Table \ref{runtimes_table}, for different values of $K$ on the GDSC drug sensitivity and MovieLens 100K datasets. Here we see that the univariate posterior models (GGGU, GEE, GEEA, GTT, GTTN, $\text{GL}^2_1$, GEG) are faster than the multivariate ones (GGG, GGGA, GGGW, GLL, GLLI); models with hierarchical priors are not noticably slower; the volume prior models are by far the slowest, due to their higher time complexity; and the Poisson models are slow for low $K$, but at higher values this is no longer true.
		
		\begin{table*}[h]
			\caption{Average runtime per iteration (in seconds) on GDSC drug sensitivity and MovieLens 100K.} 
			\label{runtimes_table}
			\centering
			\begin{tabular}{lllllllll}
				\toprule
				& \multicolumn{4}{l}{GDSC drug sensitivity} & \multicolumn{4}{l}{MovieLens 100K} \\
				\cmidrule(r){2-5} \cmidrule{6-9}
				Method & $K=5 $ & $K=10 $ & $K=20 $ & $K=50 $ & $K=5 $ & $K=10 $ & $K=20 $ & $K=50 $ \\
				\cmidrule(r){1-1} \cmidrule(r){2-2} \cmidrule(r){3-3} \cmidrule(r){4-4} \cmidrule(r){5-5} \cmidrule(r){6-6} \cmidrule(r){7-7} \cmidrule(r){8-8} \cmidrule{9-9}
				GGG & 0.14 & 0.18 & 0.29 & 0.93 & 1.04 & 1.29 & 1.86 & 5.07 \\
				GGGU & 0.02 & 0.03 & 0.07 & 0.16 & 0.48 & 0.79 & 1.49 & 3.99 \\
				GGGA & 0.14 & 0.17 & 0.28 & 0.93 & 1.03 & 1.30 & 1.88 & 5.75 \\
				GGGW & 0.13 & 0.17 & 0.26 & 0.82 & 1.27 & 1.23 & 1.84 & 5.06 \\
				GLL & 0.24 & 0.30 & 0.38 & 0.95 & 0.81 & 0.94 & 1.30 & 3.01 \\
				GLLI & 0.22 & 0.29 & 0.42 & 0.98 & 0.82 & 0.99 & 1.49 & 3.26 \\ 
				GVG & 0.42 & 1.02 & 2.58 & 22.1 & 1.22 & 2.86 & 5.94 & 40.9 \\
				\midrule
				GEE & 0.06 & 0.12 & 0.25 & 0.66 & 0.49 & 0.86 & 1.68 & 4.16 \\
				GEEA & 0.06 & 0.12 & 0.24 & 0.63 & 0.68 & 1.27 & 2.53 & 6.50 \\
				GTT & 0.06 & 0.12 & 0.25 & 0.68 & 0.70 & 1.27 & 2.34 & 6.00 \\
				GTTN & 0.07 & 0.14 & 0.29 & 0.77 & 0.71 & 1.22 & 2.55 & 6.56 \\
				$\text{GL}^2_1$ & 0.06 & 0.13 & 0.26 & 0.62 & 0.43 & 0.80 & 1.56 & 3.88 \\
				\midrule
				GVnG & 0.45 & 1.16 & 2.88 & 22.9 & 2.02 & 5.08 & 9.63 & 53.2 \\
				GEG & 0.07 & 0.12 & 0.22 & 0.61 & 0.96 & 1.30 & 1.34 & 3.51 \\
				\midrule
				PGG & 0.36 & 0.40 & 0.50 & 0.78 & 1.32 & 1.96 & 3.36 & 7.07 \\
				PGGG & 0.56 & 0.49 & 0.50 & 0.78 & 1.34 & 2.02 & 3.40 & 7.19 \\
				\midrule
				NMF-NP & 0.01 & 0.04 & 0.04 & 0.16 & 0.32 & 0.52 & 1.11 & 2.62 \\
				\bottomrule
			\end{tabular}
		\end{table*}
		
		
	\clearpage
	\section{Nested cross-validation dimensionalities}
		In the main paper we used nested cross-validation to find the best dimensionality $K$ in each fold of the cross-validation. Table \ref{crossvalidation_table} contains the average values for each model on each dataset, rounded to the nearest integer. We used these values for the noise and sparsity experiments, to give each model their own optimal starting point. The models with ARD and Wishart priors (GGGA, GGGW, GEEA) and with the volume prior (GVG, GVnG) can often leverage higher dimensionalities $K$ than the others.
		
\begin{figure}[h]
			\centering
			\includegraphics[width=\columnwidth]{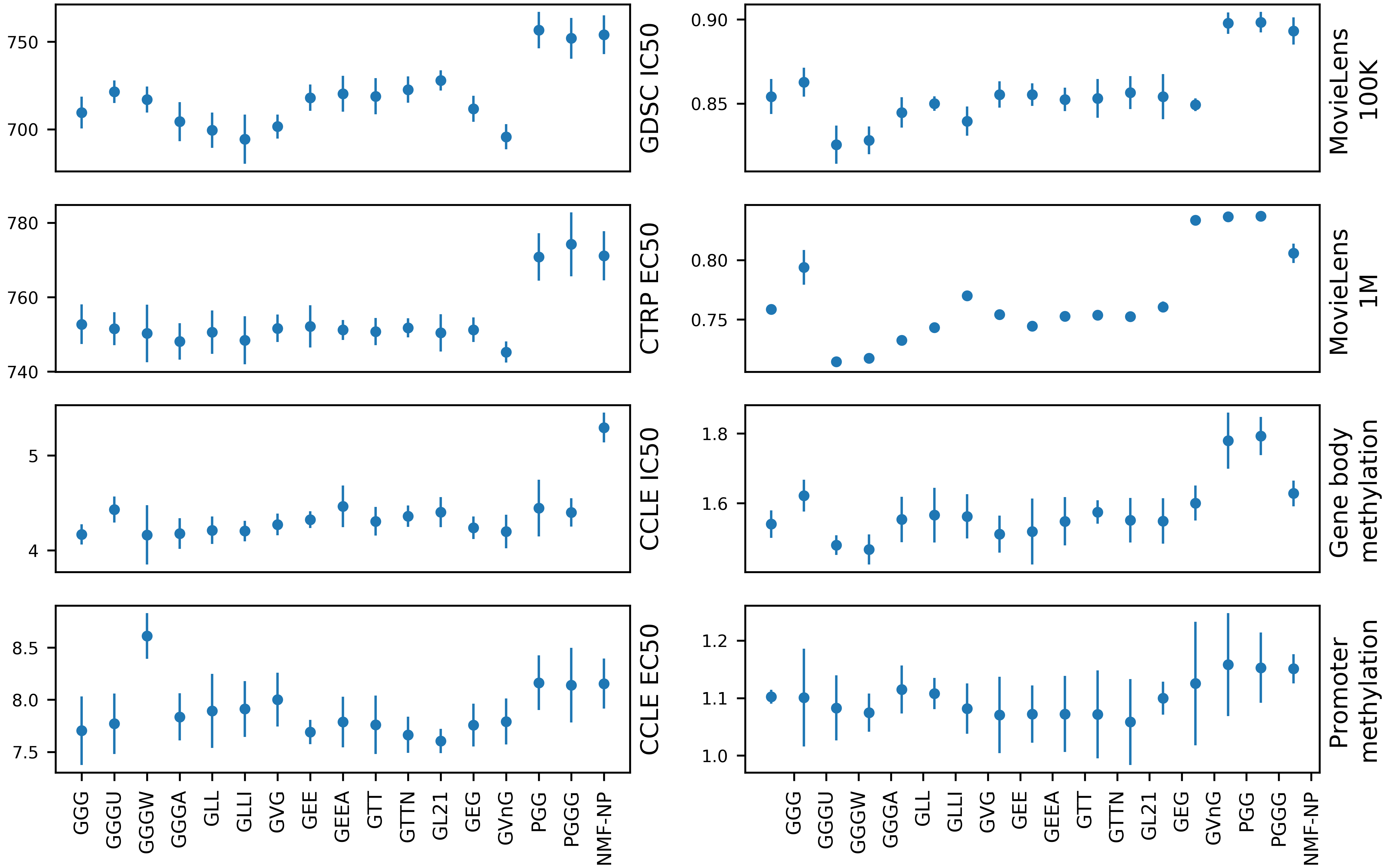}
			\caption{Average mean squared error of 5-fold nested cross-validation for the seventeen methods on the eight datasets. We also plot the standard deviation of errors across the folds.} 
			\label{crossvalidation}
		\end{figure}

		For reference, the cross-validation performances are provided in Figure \ref{crossvalidation}.
		
		\begin{table*}[h]
			\caption{Average dimensionality found in 5-fold nested cross-validation for each method on the eight datasets.} \label{crossvalidation_table}
			\centering
			\begin{tabular}{lllllllll}
				\toprule
				& \multicolumn{4}{l}{Drug sensitivity} & \multicolumn{2}{l}{MovieLens} & \multicolumn{2}{l}{Methylation} \\
				\cmidrule(r){2-5} \cmidrule{6-7} \cmidrule{8-9}
				Method & GDSC & CTRP & CCLE $IC_{50}$ & CCLE $EC_{50}$ & 100K & 1M & Gene body & Promoter \\
				\cmidrule(r){1-1} \cmidrule(r){2-2} \cmidrule(r){3-3} \cmidrule(r){4-4} \cmidrule(r){5-5} \cmidrule(r){6-6} \cmidrule(r){7-7} \cmidrule(r){8-8} \cmidrule{9-9}
				GGG & 6 & 4 & 5 & 1 & 2 & 5 & 4 & 3 \\
				GGGU & 6 & 5 & 5 & 1 & 2 & 2 & 3 & 3 \\
				GGGA & 10 & 6 & 5 & 1 & 4 & 10 & 6 & 3 \\
				GGGW & 16 & 8 & 6 & 2 & 5 & 13 & 7 & 3 \\
				GLL & 10 & 6 & 5 & 1 & 3 & 8 & 4 & 2 \\
				GLLI & 10 & 6 & 5 & 1 & 2 & 7 & 4 & 2 \\
				GVG & 10 & 5 & 6 & 2 & 3 & 3 & 4 & 4 \\
				\midrule
				GEE & 8 & 6 & 5 & 1 & 2 & 8 & 6 & 5 \\
				GEEA & 10 & 6 & 5 & 1 & 2 & 10 & 6 & 4 \\
				GTT & 9 & 6 & 5 & 1 & 2 & 8 & 5 & 4 \\
				GTTN & 8 & 6 & 5 & 1 & 2 & 8 & 5 & 4 \\
				$\text{GL}^2_1$ & 6 & 6 & 4 & 1 & 2 & 8 & 5 & 5 \\
				\midrule
				GEG & 6 & 5 & 5 & 1 & 2 & 5 & 4 & 3 \\
				GVnG & 11 & 8 & 5 & 2 & 2 & 1 & 3 & 3 \\
				\midrule
				PGG & 4 & 2 & 13 & 1 & 1 & 1 & 2 & 3 \\
				PGGG & 5 & 2 & 12 & 1 & 1 & 1 & 2 & 3 \\
				\midrule
				NMF & 4 & 2 & 1 & 1 & 1 & 3 & 2 & 2 \\
				\bottomrule
			\end{tabular}
		\end{table*}
	
	\section{Sparsity and model selection plots}
\begin{figure*}[h]
			\centering
			\includegraphics[width=0.7\textwidth]{legend_colours.png}
			\vspace{5pt}
			\includegraphics[width=0.9\textwidth]{legend.png}
			\vspace{10pt}
			\begin{minipage}{\textwidth}
				\includegraphics[width=\textwidth]{sparsity_gdsc_multiple_row.png}
				\includegraphics[width=\textwidth]{sparsity_methylation_gm_multiple_row.png}
				\includegraphics[width=\textwidth]{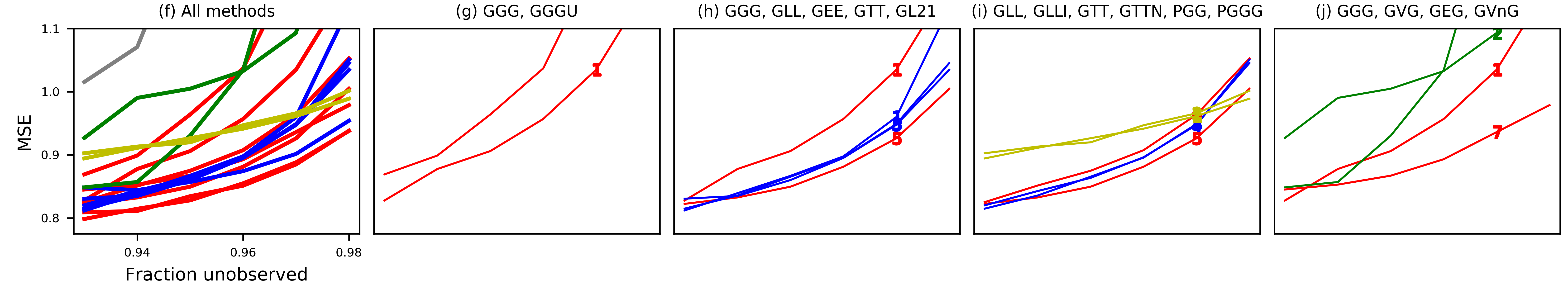}
				\caption{Sparsity experiment results on the GDSC drug sensitivity (top, a-e), gene body methylation (middle, f-j), and MovieLens 100K (bottom, k-o) datasets. We measure the predictive performance (mean squared error) on a held-out dataset for different fractions of unobserved data.}
				\label{sparsity_gm_gdsc_ml100k}
			\end{minipage}
			\begin{minipage}{\textwidth}
				\includegraphics[width=\textwidth]{model_selection_gdsc_multiple_row.png}
				\includegraphics[width=\textwidth]{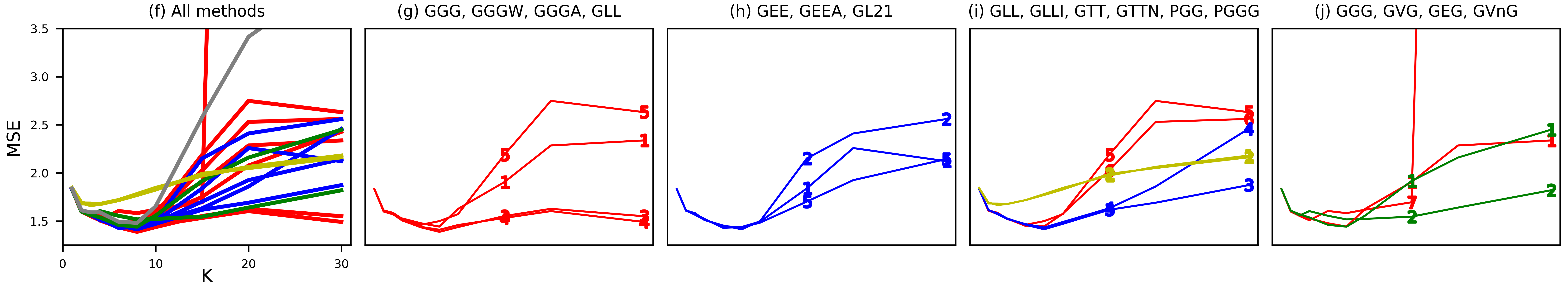}
				\includegraphics[width=\textwidth]{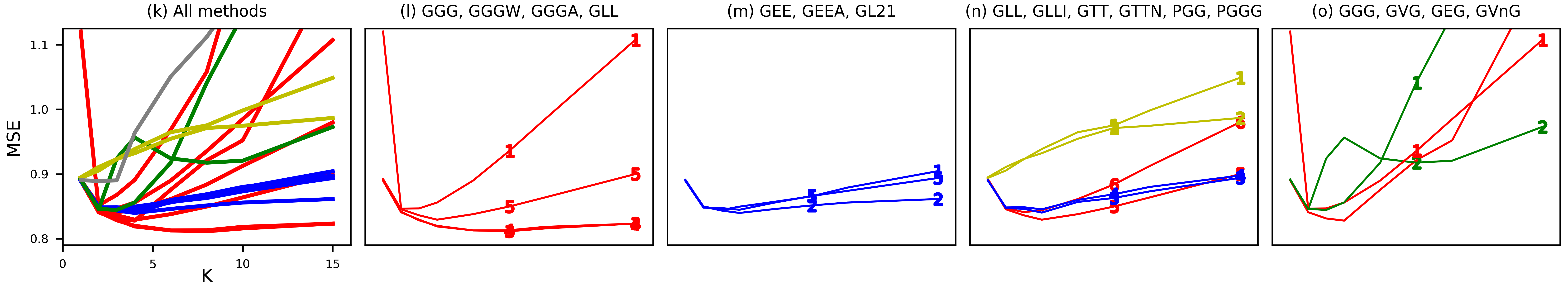}
				\caption{Model selection experiment results on the GDSC drug sensitivity (top, a-e), gene body methylation (middle, f-j), and MovieLens 100K (bottom, k-o) datasets. We measure the predictive performance (mean squared error) on a held-out dataset for different dimensionalities $K$.}
				\label{model_selection_gm_gdsc_ml100k}
			\end{minipage}
		\end{figure*}

		Although we presented results for the sparsity and model selection experiments on only a few datasets, we ran them both on the GDSC drug sensitivity, MovieLens 100K, and gene body methylation datasets. We give the results in Figures \ref{sparsity_gm_gdsc_ml100k} and \ref{model_selection_gm_gdsc_ml100k}.
	
\clearpage